\documentclass[format=acmsmall, review=false, screen=false]{acmart}

\usepackage{booktabs} % For formal tables

\usepackage{url}
\usepackage{amsmath}
\usepackage{float}
\usepackage{subfigure}
\usepackage{graphicx}
\usepackage{color}
\usepackage[noend]{algorithmic}
\usepackage{algorithm}
\usepackage{multirow}
\usepackage{verbatim}
\usepackage{amsmath}
\usepackage{amsthm}

\setcopyright{none}
\begin{document}
% Title portion. Note the short title for running heads 
\title[Spatial Classification With Limited Observations]{Spatial Classification With Limited Observations Based On Physics-Aware Structural Constraint}

\author{Arpan Man Sainju*, Wenchong He*, Zhe Jiang}
\affiliation{%
  \institution{Department of Computer Science, University of Alabama}
}
\email{asainju@crimson.ua.edu, whe11@crimson.ua.edu, zjiang@cs.ua.edu}

\author{Da Yan}
\affiliation{%
  \institution{Department of Computer Science, University of Alabama at Birmingham}
}
\email{yanda@uab.edu}

%I added this co-author below. we collaborated earier on another limited observation paper from his group
\author{Haiquan Chen}
\affiliation{%
  \institution{California State University, Sacramento}
}
\email{haiquan.chen@csus.edu}

\begin{abstract}
Spatial classification with limited feature observations has been a challenging problem in machine learning. The problem exists in applications where only a subset of sensors are deployed at certain spots or partial responses are collected in field surveys. Existing research mostly focuses on addressing incomplete or missing data, e.g., data cleaning and imputation, classification models that allow for missing feature values or model missing features as hidden variables in the EM algorithm. These methods, however, assume that incomplete feature observations only happen on a small subset of samples, and thus cannot solve problems where the vast majority of samples have missing feature observations. To address this issue, we recently proposed a new approach that incorporates physics-aware structural constraint into the model representation. Our approach assumes that a spatial contextual feature is observed for all sample locations and establishes spatial structural constraint from the underlying spatial contextual feature map. We design efficient algorithms for model parameter learning and class inference. This paper extends our recent approach by allowing feature values of samples in each class to follow a multi-modal distribution. We propose learning algorithms for the extended model with multi-modal distribution. Evaluations on real-world hydrological applications show that our approach significantly outperforms baseline methods in classification accuracy, and the multi-modal extension is more robust than our early single-modal version especially when feature distribution in training samples is multi-modal. Computational experiments show that the proposed solution is computationally efficient on large datasets.
\end{abstract}

% <concept_id>10002951.10003227.10003236.10003237</concept_id>
% <concept_desc>Information systems~Geographic information systems</concept_desc>
% <concept_id>10002951.10003227.10003351</concept_id>
% <concept_desc>Information systems~Data mining</concept_desc>

% \ccsdesc[500]{Information systems~Geographic information systems}
% \ccsdesc[500]{Information systems~Data mining}



\thanks{
Contact author: Zhe Jiang (zjiang@cs.ua.edu)\\
A.~Sainju and W.~he are co-first authors with equal contribution\\
Authors' addresses: A.~Sainju, W.~He, Z.~Jiang, Computer Science Department, the University of Alabama, Box 870290, Tuscaloosa, AL 35487, USA; D. Yan, Computer Science Department, the University of Alabama at Birmingham; H. Chen, Computer Science Department, California State University, Sacramento}

\maketitle

\renewcommand{\shortauthors}{A. Sainju and W. He et al.}

\section{Introduction}\label{sec:intro}

Given a spatial raster framework with explanatory feature layers, a spatial contextual layer (e.g., a potential field), as well as a set of training samples with class labels outside the framework, the spatial classification problem aims to learn a model that can predict a class layer~\cite{jiang2018survey,jiang2020spatial,shekhar2015spatiotemporal,du2019beyond,zhang2019unifying,wang2018learning}. We particularly focus on spatial classification with limited feature observations, i.e., only limited pixel locations in the raster framework have explanatory feature data available. For example, in earth imagery classification, the explanatory feature layers are spectral bands of earth imagery pixels; the spatial contextual layer can be elevation, and the target class layer consists of pixel classes (e.g., flood or dry). In the example, it often happens that the elevation values are available for all pixels in the framework, but only limited pixel locations have spectral data (e.g., a drone or aerial plane could not cover the entire region due to limited time during a flood disaster).

The problem is important in many applications such as flood extent mapping.
Flood extent mapping plays a crucial role in addressing grand societal challenges such as disaster management,  national water forecasting, as well as energy and food security~\cite{jiang2017spatial,Xie2018,jiang2019geographical}. For example, during Hurricane Harvey floods in 2017, first responders needed to know the spatial extent of floodwater in order to plan rescue efforts. In national water forecasting, detailed flood extent maps can be used to calibrate and validate the NOAA National Water Model~\cite{nwm}. In current practice, flood extent maps are mostly generated by flood forecasting models, whose accuracy is often unsatisfactory in a high spatial resolution~\cite{iwrss,merwade2008uncertainty}. Other ways to generate flood maps involve sending a field crew on the ground to mark down the floodwater extent on a map, but the process is both expensive and time-consuming.  A promising alternative is to utilize observation data from groundwater sensors and remote sensors on aerial planes or drones. However, sensor observations often have limited spatial coverage due to only a subset of sensors being deployed at certain spots, making it a problem of spatial classification with limited feature observations. For example, during a flood disaster, a drone can only collect spectral images in limited areas due to time limit. Note that though we use flood mapping application as an example, the problem can potentially be applied to other broad applications such as water quality monitoring~\cite{yang2010gis} in river networks.

The problem poses several unique challenges that are not well addressed by traditional classification techniques. First, there are limited feature observations on samples in the raster framework due to only a subset of sensors being deployed in certain regions. In other words, only a subset of samples have complete explanatory feature values, making it hard to predict classes for all samples. Second, among the sample pixels with complete explanatory feature values, their feature values may contain rich noise and obstacles. For example, high-resolution earth imagery often has noise: clouds and shadows. Third, the explanatory features of image pixels can be insufficient to distinguish classes (also called class confusion) due to heterogeneity. For instance, pixels of tree canopies overlaying flood water have the same spectral features as trees in dry areas, yet their classes are different. Finally, the number of pixel locations can be very large for high-resolution data (e.g., hundreds of millions of locations in one city), requiring scalable algorithms.

Over the years, various techniques have been developed to address missing feature observations (or incomplete data) in classification~\cite{garcia2010pattern}. Existing methods can be categorized into data cleaning or imputation, utilizing classification models that allow for missing feature values, and modeling missing features as hidden variables in the EM (Expectation-Maximization) algorithm. Data cleaning will remove samples that miss critical feature values. Data imputation focuses on filling in missing values either by statistical methods~\cite{little2019statistical} (e.g., mean feature values from observed samples) or by prediction models (e.g., regression) based on observed samples~\cite{schafer1997analysis,batista2002study,yoon1999training,bengio1996recurrent,rubin2004multiple}. Another approach focuses on classification models and algorithms that allow for missing feature values in learning and prediction without data imputation. For example, a decision tree model allows for samples with missing features in learning and classification~\cite{quinlan2014c4,quinlan1989unknown,webb1998problem}. During learning, for a missing feature value in a sample, a probability weight is assigned to each potential feature value based on its frequency in observed samples. During classification, a decision tree can explore all possible tree traversal paths for samples with missing features and select the final class prediction with the highest probability. Similarly, there are some other models that have been extended to allow for missing feature values, such as neural network ensembles~\cite{jiang2005classification}, and support vector machine~\cite{chechik2007max,smola2005kernel,pelckmans2005handling}. The last category is to model missing feature values as hidden variables and use the EM (Expectation-Maximization) algorithm for effective learning and inference~\cite{mclachlan2007algorithm,ghahramani1994supervised,ghahramani1995learning,williams2007classification}. Specifically, the joint distribution of all samples' features (both observed and missing features) can be represented by a mixture model with fixed but yet unknown parameters. In the EM algorithm, we can use initialized parameters and observed features to estimate the posterior distribution of hidden variables (missing features), and then further update the parameters for the next iteration. 
However, all these existing methods assume that incomplete feature observations only happen on a small subset of samples, and thus cannot be effectively applied to our problem where the vast majority of samples have missing features (i.e., limited feature observations). 

To fill this gap, we recently proposed a new approach that incorporates physics-aware structural constraints into model representation~\cite{sainju2020spatial}. Our approach assumes that a spatial contextual feature is fully observed for every sample location, and establishes spatial structural constraints from the spatial contextual feature map. We design efficient algorithms for model parameter learning and class inference and conduct experimental evaluations to validate the effectiveness and efficiency of the proposed approach against existing works. This journal paper extends our recent approach by allowing for feature values of samples in each class to follow a multi-modal distribution. We also propose the parameter learning algorithms for the extended model. In summary, the paper makes the following contributions:

\begin{itemize}
    \item We propose an approach that utilizes physics-aware spatial structural constraints to handle limited feature observations in spatial classification. 
    \item We design efficient algorithms for model parameter learning and class inference. 
    \item We extend the model by allowing feature values of samples in each class to follow a multi-modal distribution.
    \item We evaluated the proposed model on two real-world hydrological datasets. Results show that our approach significantly outperforms several baseline methods in classification accuracy, and the new multi-modal solution is more robust than the previous single-modal version especially when feature distribution in training samples is multi-modal.
    \item Computation experiments show that the proposed solution is scalable to a large data volume.
\end{itemize}

\section{Problem Statement}\label{sec:prob}
\subsection{Preliminaries}
Here we define several basic concepts that are used in the problem formulation.

A \emph{spatial raster framework} is a tessellation of a two-dimensional plane into a regular grid of $N$ cells. A spatial neighborhood relationship exists between cells based on cell adjacency. The framework can contain $m$ non-spatial explanatory feature layers (e.g., spectral bands in earth imagery), one potential field layer (e.g., elevation), and one class layer (e.g., \emph{flood} or \emph{dry}).

Each cell in a raster framework is a \emph{spatial data sample}, denoted by $\mathbf{s_n}=(\mathbf{x}_n, \phi_n, y_n)$, where $n\in \mathbb{N}, 1\leq n \leq N$, $\mathbf{x}_n\in \mathbb{R}^{m\times 1}$ is a vector of $m$ non-spatial explanatory feature values with each element corresponding to one feature layer, $\phi_n\in\mathbb{R}$ is a cell's potential field value, and $y_n\in \{0,1\}$ is a binary class label. 

A raster framework with all samples is denoted by $\mathcal{F}=\{\mathbf{s_n}|n\in \mathbb{N}, 1\leq n \leq N\}$, non-spatial explanatory feature matrix of all samples  are denoted by $\mathbf{X}=[\mathbf{x}_1,...,\mathbf{x}_N]^T$, the potential field vector is denoted by $\boldsymbol{\Phi}=[\phi_1,...,\phi_N]^T$, and the class vector is denoted by $\mathbf{Y}=[y_1,...,y_N]^T$. 

In a raster framework, it may happen that only a limited number of samples have non-spatial explanatory features being observed. We define $\mathcal{O}$ as the set of indices for these fully observed samples. Samples with fully observed (complete) explanatory features are denoted by $\{\mathbf{x}_n|n\in\mathcal{O}\}$. Their feature matrix is denoted by $\mathbf{X_o}$. 

\subsection{Problem Definition}

\begin{figure}[h]
    \centering
    \subfigure[Spatial potential field (elevation) ]{\includegraphics[width=1in,height=1in]{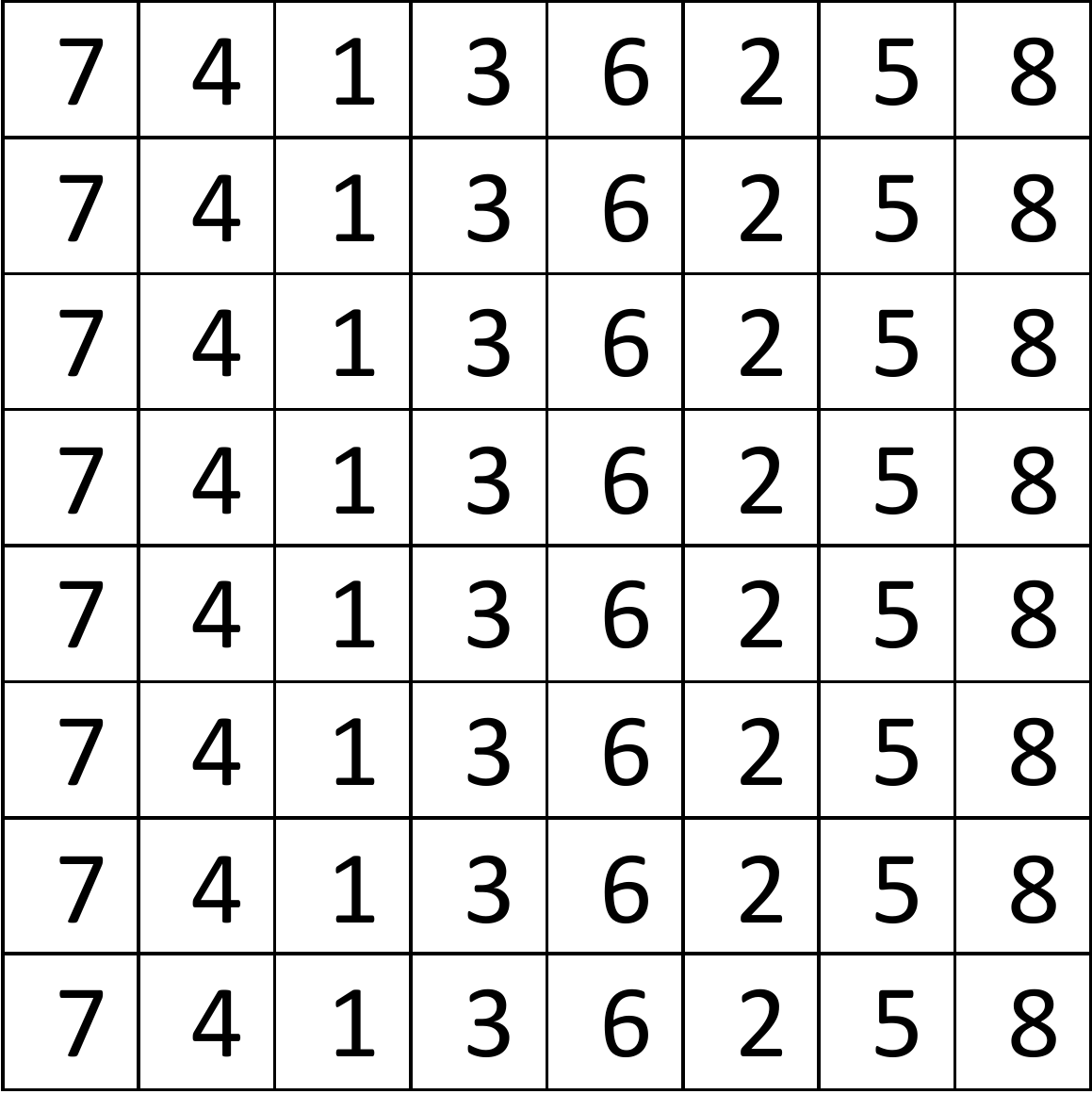}}
    \hspace{2mm}
    \subfigure[Partially observed non-spatial feature values]{\includegraphics[width=1in,height=1in]{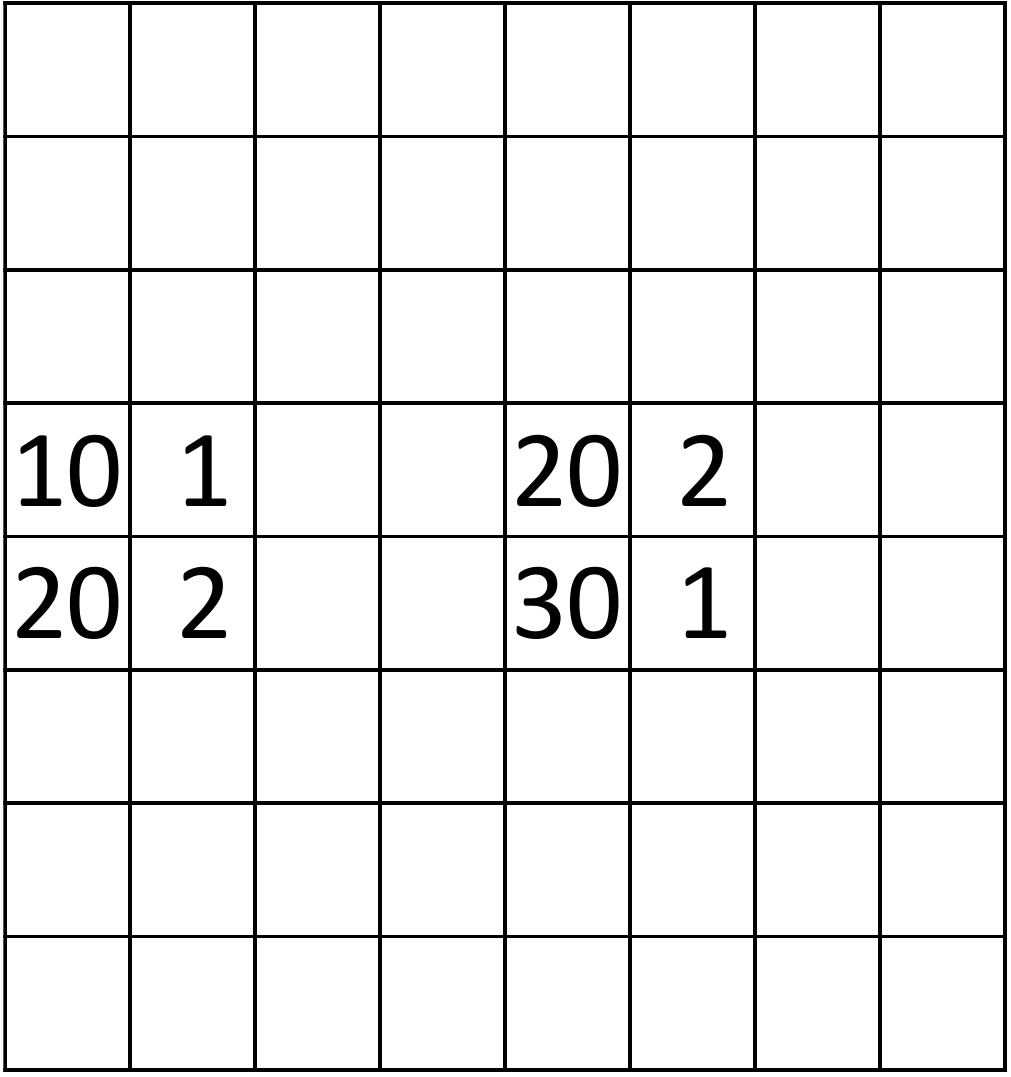}}\hspace{2mm}
    \subfigure[Ground truth classes (green for dry, orange for flood)]{\includegraphics[width=1in,height=1in]{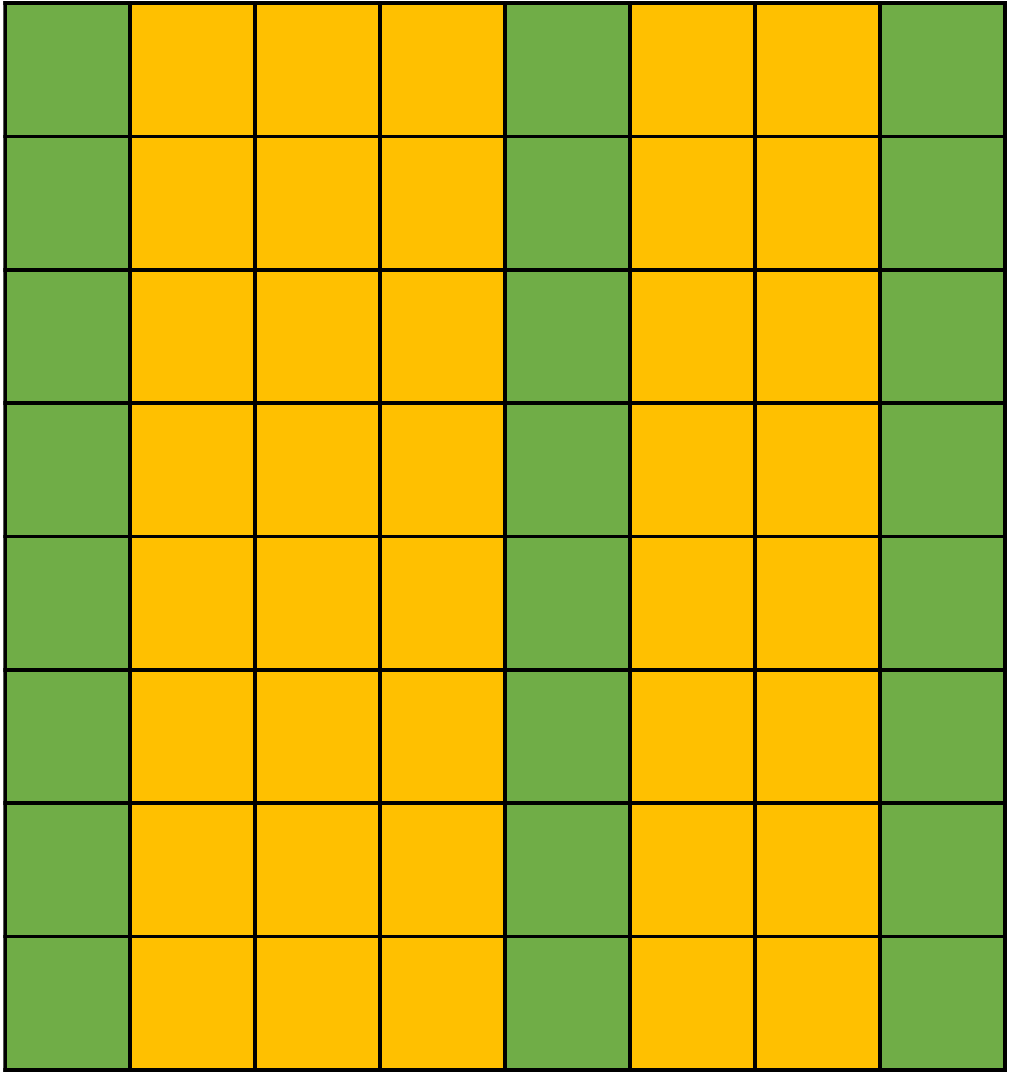}}
    \caption{An illustration problem example}
    \label{fig:ToyExample}
\end{figure}

Given a raster framework with the explanatory features of a limited number of samples $\mathbf{X}_o$, the potential field layer of all samples in the framework $\boldsymbol{\Phi}=[\phi_1,...,\phi_N]^T$, and a set of training samples with class labels outside the framework, the spatial classification problem aims to learn a classifier $f$ to predict the class layer $\mathbf{Y}=f(\mathbf{X}_o,\boldsymbol{\Phi})$. 
For example, in earth imagery classification, the explanatory feature layers are spectral bands of earth imagery pixels; the spatial contextual layer can be elevation, and the target class layer consists of pixel classes (e.g., flood or dry). We assume that the elevation values are available for all pixels in the framework (in practice, elevation values do not change over time and thus can be collected all at once) but only limited pixel locations have spectral data (e.g., a drone or aerial plane could not cover the entire region due to time limit). Figure~\ref{fig:ToyExample} shows a toy example of a raster framework that consists of sixty-four samples with a one-dimensional explanatory feature and a potential field layer. There are only eight samples with observed explanatory features (four non-empty cells in Figure~\ref{fig:ToyExample}(b)). The goal is to learn a model that can predict the class layer in Figure~\ref{fig:ToyExample}(c).

\section{Approach}\label{sec:app}
In this section, we introduce our proposed approach. We start with physics-aware structural constraints and then introduce our probabilistic model and its learning and inference algorithms. We will introduce our approach in the context of the flood mapping application, but the proposed method can be potentially generalized to other applications such as material science~\cite{wales1998archetypal,wales2003energy} and biochemistry~\cite{edelsbrunner2005geometry,gunther2014characterizing}.

\subsection{Physics-Aware Structural Constraint}
\begin{figure}[h]
    \centering
    \subfigure[Eight consecutive sample locations in one dimensional space]{\includegraphics[width=1.5in]{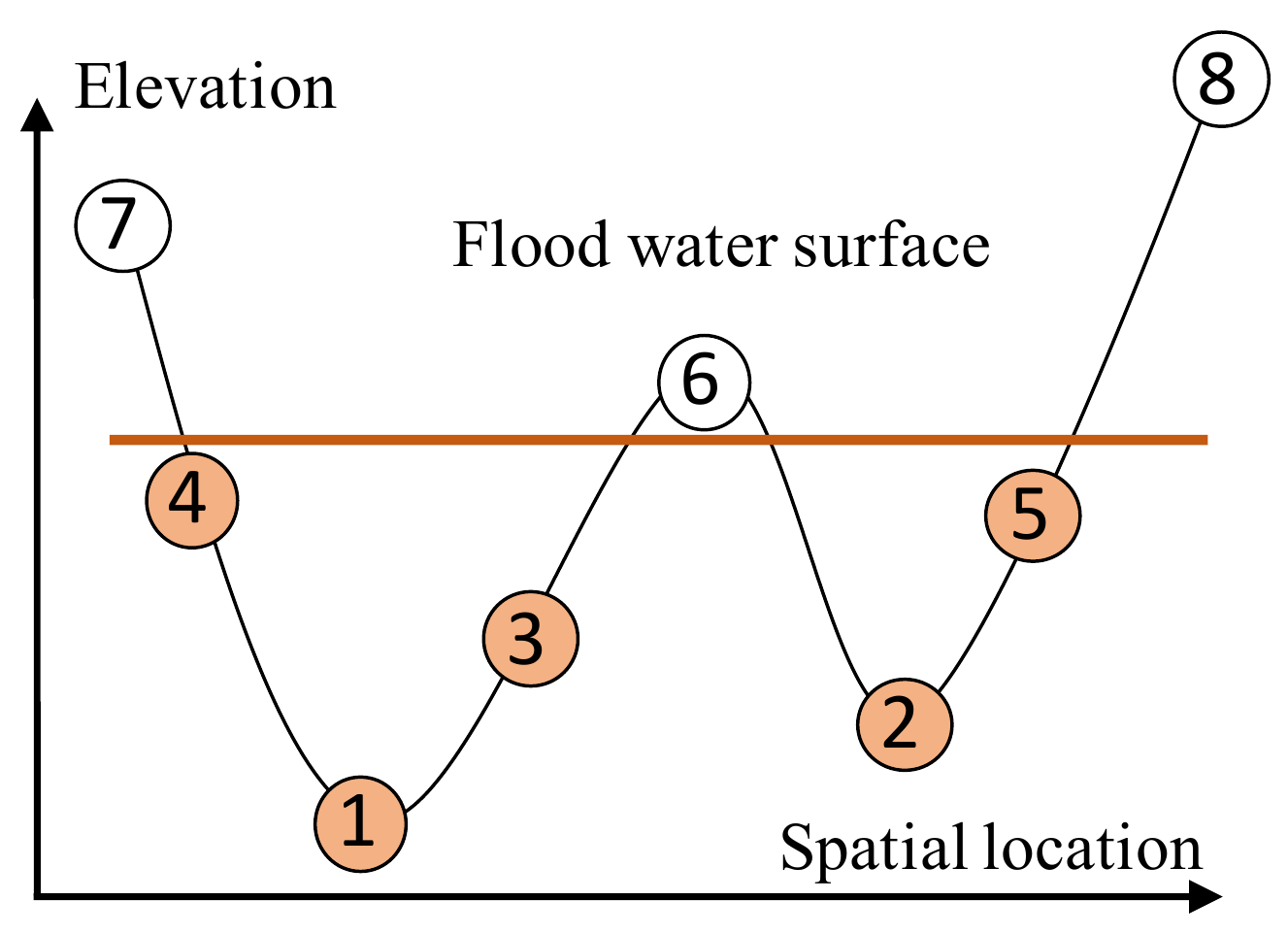}}\hspace{1mm}
    \subfigure[Partial order constraint in a reverse tree]{\includegraphics[width=1in]{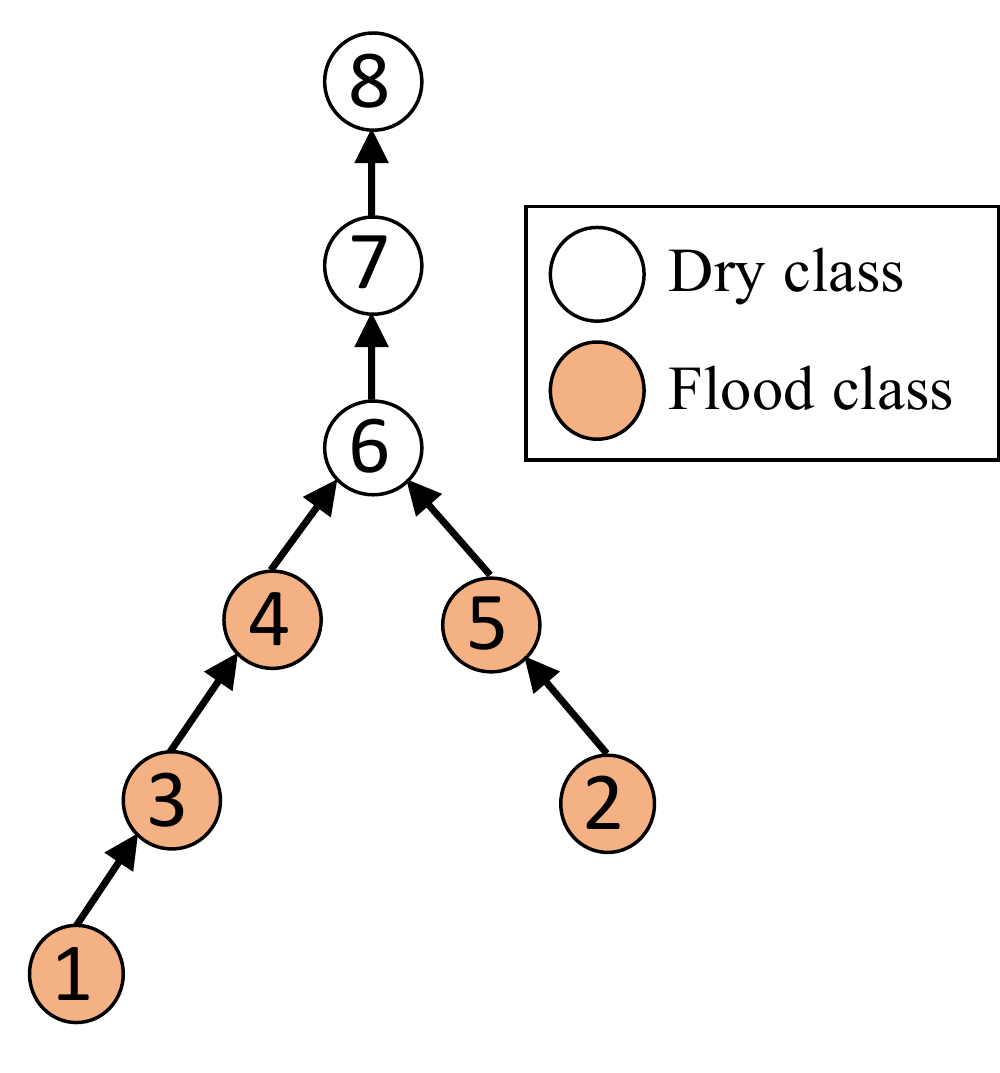}}\hspace{1mm}
    \caption{Illustration of partial order class dependency}
    \label{fig:tree}
\end{figure}

The main idea of our proposed approach is to establish a spatial dependency structure of sample class labels based on the physical constraint from the spatial potential field layers (e.g., water flow directions based on elevation). An illustration is provided in Figure~\ref{fig:tree}. Figure~\ref{fig:tree}(a) shows the elevation values of eight pixels in one dimensional space (e.g., pixels on a row in Figure~\ref{fig:ToyExample}). Due to gravity, water flows from high locations to nearby lower locations. If location $4$ is flooded, locations $1$ and $3$ must also be flooded. Such a dependency structure can be established based on the topology of the potential field surface (e.g., elevation). Figure~\ref{fig:tree}(b) shows a directed tree structure that captures the flow dependency structure. If any node is \emph{flood}, then all sub-tree nodes must also be \emph{flood} due to gravity. The structure is also called \emph{split tree} in topology~\cite{carr2003computing,edelsbrunner2010computational}, where a node represents a vertex on a mesh surface (spatial potential field) and an edge represents the topological relationships between vertices. We can efficiently construct the tree structure from a potential field map following the topological order of pixels based on the union-find operator (its time complexity is $O(N\log N)$)~\cite{carr2003computing}. We omit details due to space limit. It is worth noting that although our illustrative example in Figure~\ref{fig:tree} is in one-dimensional space for simplicity, the structure is readily applicable to two-dimensional space~\cite{jiangkdd2019}. We can create a single tree structure for the entire elevation map in Figure~\ref{fig:ToyExample}(a).

\subsection{Model Probabilistic Formulation}
\begin{figure}[h]
    \centering
    \includegraphics[width=1.5in]{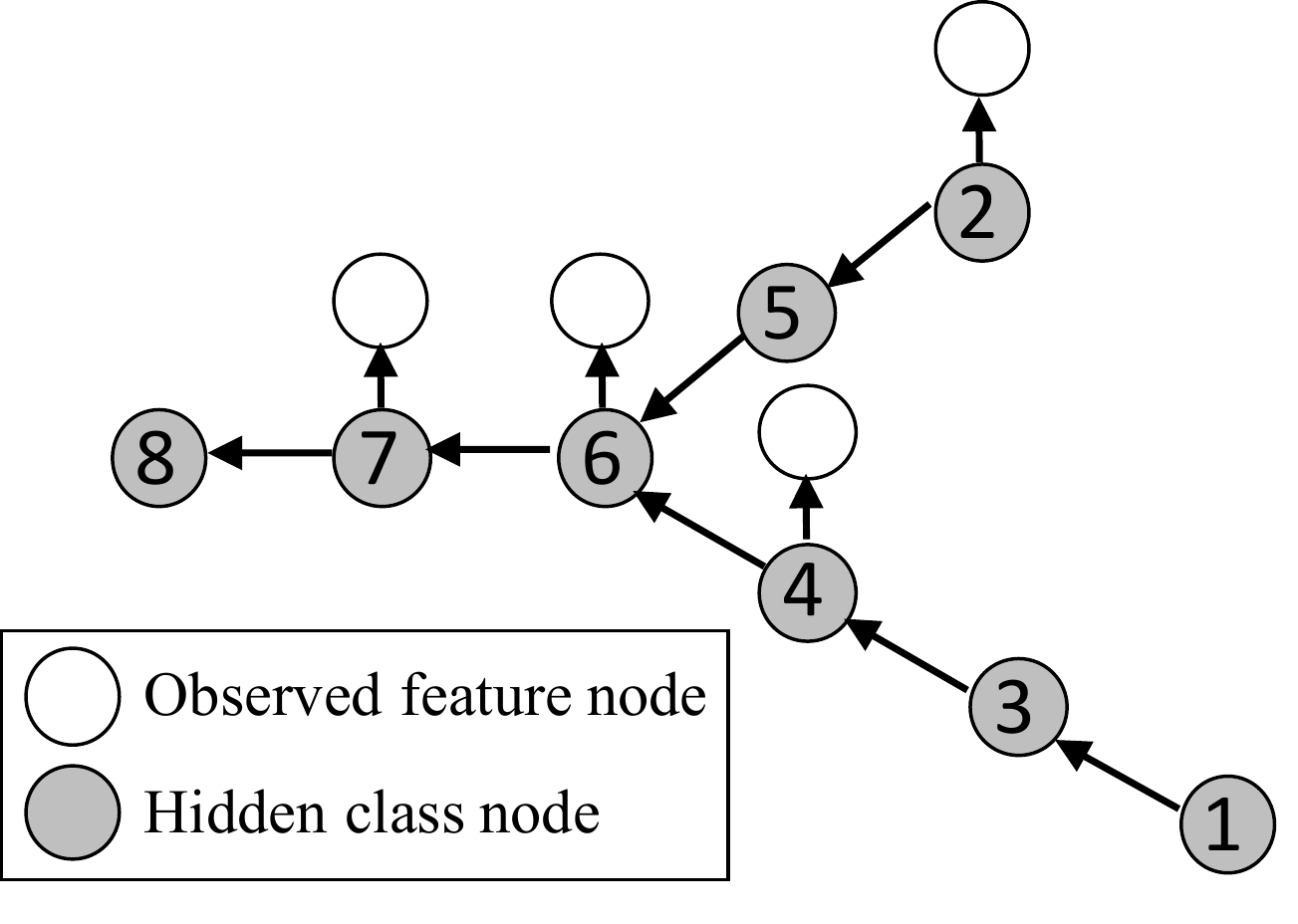}
    \caption{Illustration of model structure}
    \label{fig:hmtrt}
\end{figure}

Now we introduce our approach that integrates physics-aware structural constraint into the probabilistic model formulation to handle limited feature observations.  Figure~\ref{fig:hmtrt} illustrates the overall model structure. It consists of two layers: a hidden class layer with unknown sample classes ($y_n$) and an observation layer with limited sample feature vectors ($\mathbf{x}_n$). Each node corresponds to a spatial data sample (raster cell). Edge directions show a probabilistic conditional dependency structure based on physical constraint. %Specifically, the model assumes that feature vectors of different samples are conditionally independent with each other given their classes, and sample classes follow a partial order dependency in a reverse tree structure.

The joint distribution of all observed samples' features and classes can be expressed as Equation~\ref{eq:joint}, where ${\mathcal{P}_n}$ is the set of parent samples of the $n$-th sample in the dependency tree, and $y_{k\in\mathcal{P}_n}\equiv\{y_k|k\in\mathcal{P}_n\}$ is the set of class nodes corresponding to parents of the $n$-th sample. For a leaf node $n$, ${\mathcal{P}_n}=\emptyset$, and $P(y_n|y_{k\in\mathcal{P}_n})=P(y_n)$.
\begin{equation}\label{eq:joint}\footnotesize
    P(\mathbf\mathbf{X}_o,\mathbf{Y})= P(\mathbf\mathbf{X}_o|\mathbf{Y})P(\mathbf{Y}) = \prod_{n\in \mathcal{O}} P(\mathbf{x}_{n}|y_{n}) \prod_{n=1}^NP(y_n|y_{k\in\mathcal{P}_n})
\end{equation}

The conditional probability of sample feature vector given its class can be assumed i.i.d. Gaussian for simplicity, as shown in Equation~\ref{eq:featureclassprob}, where $\boldsymbol{\mu}_{y_n}$ and $\boldsymbol{\Sigma}_{y_n}$ are the mean and covariance matrix of feature vector $\mathbf{x}_n$ for class $y_n$ ($y_n=0,1$). It is worth noting that $P(\mathbf{x}_n|y_n)$ could be more general than i.i.d. Gaussian.
\begin{equation}\label{eq:featureclassprob}
    P(\mathbf{x}_n|y_n)\sim \mathcal{N}(\boldsymbol{\mu}_{y_n},\boldsymbol{\Sigma}_{y_n})
\end{equation}

Class transitional probability follows the partial order constraint. For example, due to gravity, if any parent's class is \emph{dry}, the child's class must be \emph{dry}; if all parents' classes are \emph{flood}, then the child's class has a high probability of being \emph{flood}. Consider \emph{flood} as the positive class (class  $1$) and \emph{dry} as the negative class (class  $0$), then the previous reasoning shows that class transitional probability is actually conditioned on the product of parent classes $y_{\mathcal{P}_n}\equiv\prod_{k\in\mathcal{P}_n}y_k$. Table~\ref{tab:transitionprob} shows two parameters ($\pi$ and $\rho$) for class transitional probability and class prior probability.

\begin{table}[h]
\caption{Class transition probability and prior probability}
\label{tab:transitionprob}
\begin{tabular}{|c|c|c|}\hline
$P(y_n|y_{\mathcal{P}_n})$ & $y_{\mathcal{P}_n}=0$ & $y_{\mathcal{P}_n}=1$\\ \hline
$y_n=0$ & $1$ & $1-\rho$\\ \hline
$y_n=1$ & $0$ & $\rho$ \\ \hline
\end{tabular}\hspace{2mm}
\begin{tabular}{|c|c|c|}\hline
 & $P(y_n)$\\ \hline
$y_n=0$ & $1-\pi$\\ \hline
$y_n=1$ & $\pi$ \\ \hline
\end{tabular}
\end{table}

\subsection{Model Parameter Learning and Class Inference}
Our model parameters include the mean and covariance matrix of sample features in each class, the prior probability of leaf node classes, and class transition probability for non-leaf nodes. We denote the entire set of parameters as $\boldsymbol{\Theta}=\{\rho, \pi, \boldsymbol{\mu}_c, \boldsymbol{\Sigma}_c|c=0,1 \}$. Learning the set of parameters poses two major challenges: first, Equation~\ref{eq:joint} contains both unknown parameters and hidden class variables $\mathbf{Y}=[y_1,...,y_N]^T$ that are non-i.i.d.; second, the number of samples ($N$) can be huge (e.g., millions of pixels).

To address these challenges, we propose to use the expectation-maximization (EM) algorithm together with message (belief) propagation. The main idea of the EM algorithm is to first initialize a parameter setting, and compute the posterior expectation of log-likelihood (Equation~\ref{eq:joint}) on hidden class variables (E-step). The posterior expectation is a function of unknown parameters. Thus, we can update parameters by maximizing the posterior expectation (M-step). The two steps can repeat iteratively until the parameter values converge. One remaining issue is the calculation of posterior expectation of log-likelihood on hidden class variables. This requires to compute the marginal posterior distribution of $P(y_n, y_{k\in \mathcal{P}_n}|\mathbf{O},\boldsymbol{\Theta_0})$ and $P(y_n)$ for each node $\mathbf{s}_n$. This is very challenging due to the high dimensionality of $\mathbf{Y}$. To address this challenge, we use message propagation. Message propagation is based on the sum and product algorithm~\cite{kschischang2001factor,ronen1995parameter}. Propagation of message along nodes in a graph (or tree) is equivalent to marginalizing out node variables in the overall joint distribution in Equation~\ref{eq:joint}. Due to the space limit, we only show the major steps in the following discussion. More details of the theoretical proof are in the appendix.

\begin{figure}[h]
\centering
\subfigure[From leaves to root]{%
      \includegraphics[width=1.5in]{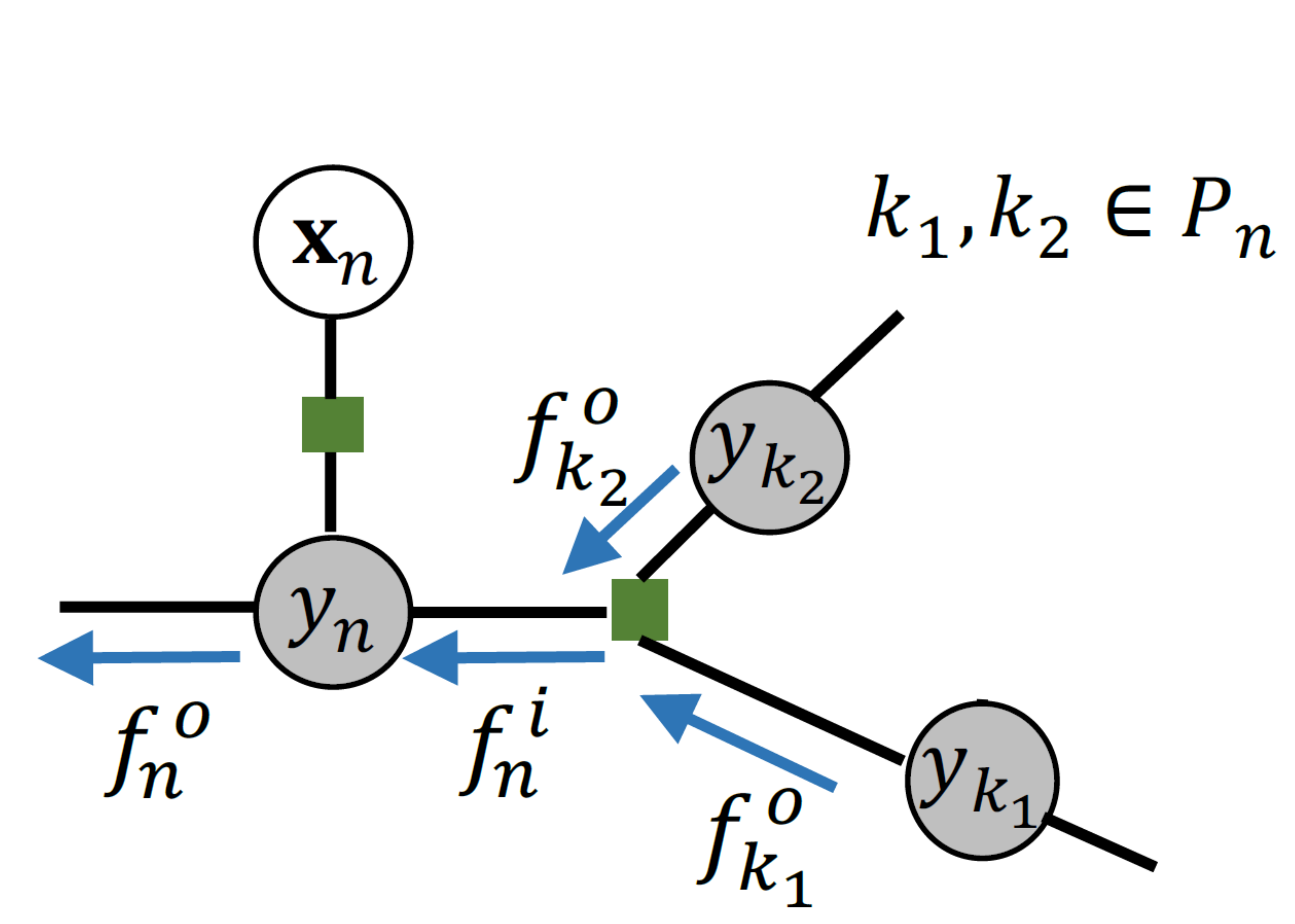}
}
\subfigure[From root to leaves]{%
      \includegraphics[width=1.28in]{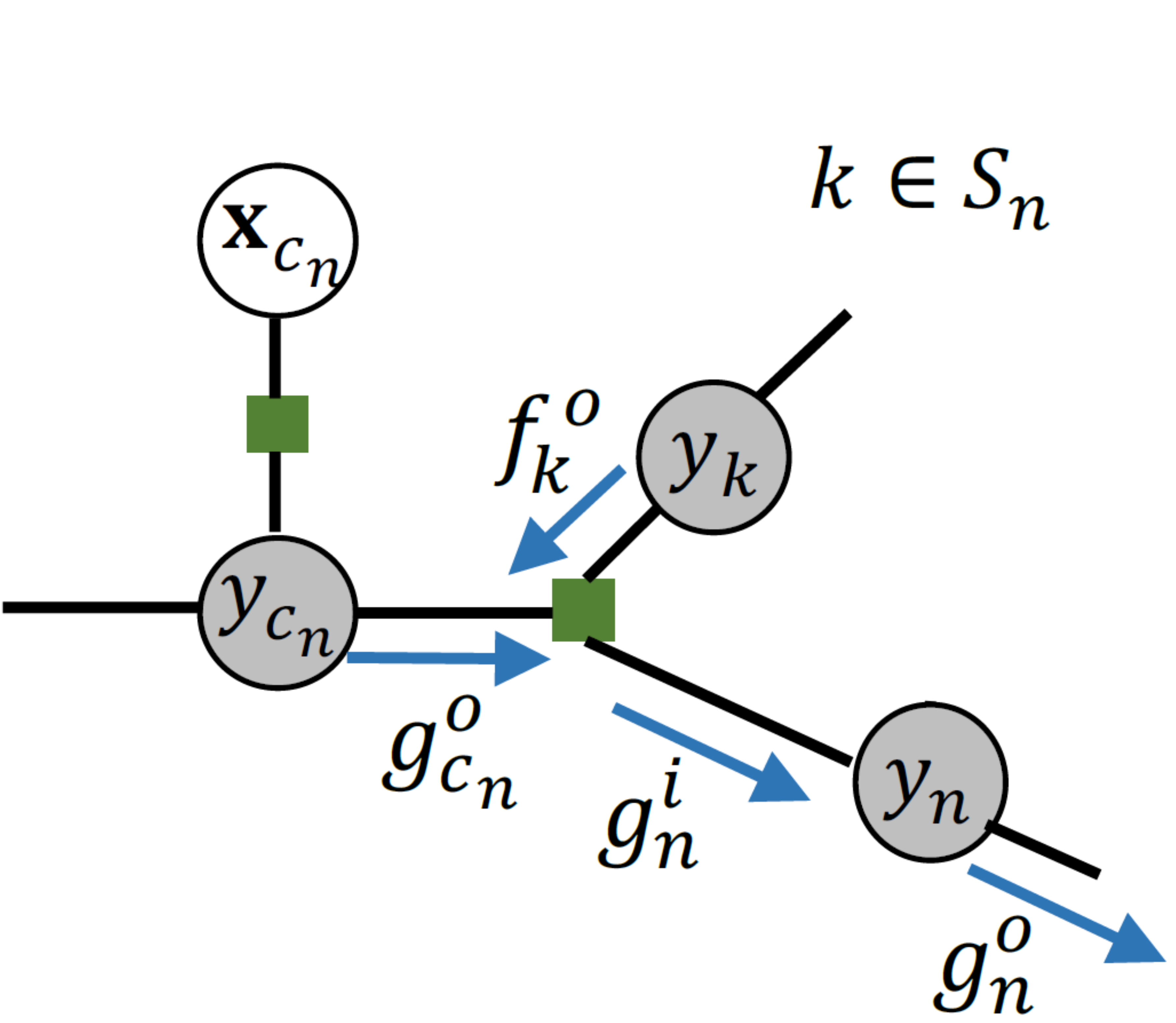}
}
\caption{Illustration of message propagation in split tree}
\label{fig:message}
\end{figure}

The message passing process is illustrated in Figure~\ref{fig:message}. 
Specifically, forward message propagation from leaves to root is based on Equation~\ref{eq:hmtforwardin} and Equation~\ref{eq:hmtforwardout}, where $f_n^i(y_n)$ and $f_n^o(y_n)$ are the incoming message into and outgoing message from a hidden class node $y_n$ respectively.
Backward message propagation from root to leaves also follows a recursive process, as shown in Equation~\ref{eq:hmtbackwardin} and Equation~\ref{eq:hmtbackwardout}, where $g_n^i(y_n)$ and $g_n^o(y_n)$ are the incoming and outgoing messages for class node $y_n$ respectively. 
For those samples without feature vector $x_n$, the outgoing forward and backward messages are the same as incoming forward and backward messages respectively, because we do not consider the feature probability for those samples. 

\begin{equation}\label{eq:hmtforwardin}\footnotesize
    f_n^i(y_n) =
            \begin{cases}
                \quad\quad\quad\quad\quad\quad P(y_n) & \text{if } y_n \text{ is leaf}\\
                \sum\limits_{y_{k\in \mathcal{P}_n}}P(y_n|y_{k\in \mathcal{P}_n})\prod\limits_{k\in \mathcal{P}_n}f_k^o(y_k) & \text{otherwise}
            \end{cases}
\end{equation}
\begin{equation}\label{eq:hmtforwardout}\footnotesize
    f_n^o(y_n) = 
            \begin{cases}
                f_n^i(y_n)P(\mathbf{x}_n|y_n) & \text{if } n \in \mathcal{O} \\
                f_n^i(y_n) &
                 \text{otherwise } 
            \end{cases}
\end{equation}

\begin{equation}\label{eq:hmtbackwardin}\footnotesize
g_n^i(y_n) =
    \begin{cases}
        \quad\quad\quad\quad\quad\quad1 & \text{if } y_n \text{ is root}\\
        \sum\limits_{y_{c_n},y_{k\in \mathcal{S}_n}}g_{c_n}^oP(y_{c_n}|y_n)\prod\limits_{k\in \mathcal{S}_n}f_k^o(y_k) & \text{otherwise}
    \end{cases}    
\end{equation}
\begin{equation}\label{eq:hmtbackwardout}\footnotesize
    g_n^o(y_n) =
            \begin{cases}
                g_n^i(y_n)P(\mathbf{x}_n|y_n) & \text{if } n \in \mathcal{O}\\
                g_n^i(y_n) &
                \text{otherwise }
            \end{cases}
\end{equation}

After both forward and backward message propagation, we can compute the marginal posterior distribution of hidden class variables based on the equations below, where $P^\prime$ is the unnormalized marginal distribution. 
\begin{equation}\label{eq:unnormalizedMarginY}\footnotesize
P^\prime(y_n|\mathbf{X}_o,\boldsymbol{\Theta_0}) =
            \begin{cases}
                f_n^i(y_n) g_n^i(y_n) P(\mathbf{x}_n|y_n)  &  \text{if } n \in \mathcal{O}\\
                f_n^i(y_n) g_n^i(y_n) &
                \text{otherwise }
            \end{cases}
\end{equation}        

\begin{equation}\label{eq:unnormalizedMarginYYp}\footnotesize
P^\prime(y_n, y_{k\in \mathcal{P}_n}|\mathbf{X}_o,\boldsymbol{\Theta_0})= \prod\limits_{k\in \mathcal{P}_n}f_k^o(y_k) g_n^o(y_n)    
\end{equation}

\begin{equation}\label{eq:normalizedMarginY}\footnotesize
P(y_n|\mathbf{X}_o,\boldsymbol{\Theta_0})\leftarrow \frac{P^\prime(y_n|\mathbf{X}_o,\boldsymbol{\Theta_0})}{\sum\limits_{y_n}P^\prime(y_n|\mathbf{X}_o,\boldsymbol{\Theta_0})}    
\end{equation}
\begin{equation}\label{eq:normalizedMarginYYp}\footnotesize
P(y_n, y_{k\in \mathcal{P}_n}|\mathbf{X}_o,\boldsymbol{\Theta_0})\leftarrow\frac{P^\prime(y_n, y_{k\in \mathcal{P}_n}|\mathbf{X}_o,\boldsymbol{\Theta_0})}{\sum\limits_{y_n,y_{k\in \mathcal{P}_n}}P^\prime(y_n, y_{k\in \mathcal{P}_n}|\mathbf{X}_o,\boldsymbol{\Theta_0})}    
\end{equation}

After computing the marginal posterior distribution, we can update model parameters by maximizing the posterior expectation of log-likelihood (the maximization or M step in EM), as shown by equations below. 
\begin{equation}\label{eq:updaterho}\footnotesize
   \rho = \frac{\sum\limits_{n|\mathcal{P}_n\neq\emptyset}{\sum\limits_{y_n}\sum\limits_{y_{\mathcal{P}_n}}{y_{\mathcal{P}_n}(1 - y_n)P(y_n, y_{\mathcal{P}_n}|\mathbf{X}, \boldsymbol{\Theta_0})}}} 
 {\sum\limits_{n|\mathcal{P}_n\neq\emptyset}{\sum\limits_{y_n}\sum\limits_{y_{\mathcal{P}_n}}{y_{\mathcal{P}_n}P(y_n, y_{\mathcal{P}_n}|\mathbf{X}, \boldsymbol{\Theta_0})}}} 
\end{equation}
%parameter update

\begin{equation}\label{eq:updatepi}\footnotesize
  \pi = \frac{\sum\limits_{n|\mathcal{P}_n=\emptyset}{\sum\limits_{y_n}{ y_n P(y_n|\mathbf{X}, \boldsymbol{\Theta_0})}}} {\sum\limits_{n|\mathcal{P}_n=\emptyset}{\sum\limits_{y_n}{P(y_n|\mathbf{X}, \boldsymbol{\Theta_0})}}}  
\end{equation}

\begin{equation}\label{eq:updatemu}\footnotesize
    \mu_c = \frac{\sum\limits_{n \in \mathcal{O}} \mathbf{x}_n P(y_n = c|\mathbf{X},\boldsymbol{\Theta_0})} {\sum\limits_{n \in \mathcal{O}}  P(y_n = c|\mathbf{X},\boldsymbol{\Theta_0})},c = {0, 1}
\end{equation}

\begin{equation}\label{eq:updatesigma}\footnotesize
    \Sigma_c = \frac{\sum\limits_{n \in \mathcal{O}} (\mathbf{x}_n - \boldsymbol{\mu}_c) (\mathbf{x}_n - \boldsymbol{\mu}_c)^T P(y_n = c|\mathbf{X},\boldsymbol{\Theta_0})} {\sum\limits_{n \in \mathcal{O}} P(y_n = c|\mathbf{X},\boldsymbol{\Theta_0})}, c = {0, 1}
\end{equation}

After learning model parameters, we can infer hidden class variables by maximizing the overall probability. A naive approach that enumerates all combinations of class assignments is infeasible due to the exponential cost. We use a dynamic programming-based method called \emph{max-sum}~\cite{rabiner1989tutorial}. The process is similar to the sum and product algorithm above. The main difference is that instead of using a \emph{sum} operation, we need to use a \emph{max} operation in message propagation, and also memorize the optimal variable values. We omit the details due to space limit. 

\subsection{Intuitions on How the Model Works}
The main intuition behind how our model handles limited observations is that the model can capture physical constraints between sample classes. The spatial structural constraints are derived from the potential field layer that is fully observed on the entire raster framework, regardless of whether non-spatial features are available or not. The topological structure in a split tree is consistent with the physical law of water flow directions on a topographic surface based on gravity. In this sense, even though many samples in the raster framework do not have non-spatial explanatory features observed, we can still infer their classes based on information from the pixels in the upstream or downstream locations.  

Another potential question is how our model can effectively learn parameters given very limited observations. This question can be answered from the perspective of how model learning works. The major task of model learning is to effectively update parameters of $P(\mathbf{x}_n|y_n)$ for observed pixels in the test region, so that we can infer the posterior class probabilities on these pixels and further infer hidden classes on other pixels. As long as the training samples could give the model a reasonable initial estimate of posterior class probabilities on the observed pixels (e.g., truly dry pixels having a higher probability of being dry), the update of parameters should be effective. This is because that parameter updates are largely weighted average of the sample mean and covariance matrices on fully observed pixels. The corresponding weights are the posterior class probability of observed samples. 

\subsection{Extension to Multi-modal Feature Distribution}
This subsection introduces an extension to our proposed model. In the conference version, the joint distribution of all sample features and classes follow a conditional independence assumption based on the tree structure derived from physical constraint (Equation~\ref{eq:joint}). Sample feature distribution in each class is assumed i.i.d. Gaussian (Equation~\ref{eq:featureclassprob}). In real-world datasets, the actual sample feature distribution in each class can be multi-modal, violating the earlier assumption on a single-modal Gaussian distribution. To account for this observation, we extend our model to allow for multi-modal sample feature distribution in each class. Specifically, we assume that sample features follow an i.i.d. mixture of Gaussian distribution. 

For the extended model, the joint distribution of all observed samples' features and classes can be expressed the same way as Equation~\ref{eq:joint}. The joint probability can be decomposed into local factors, i.e., the conditional distribution of features in each class and class transitional probability. The assumption on class transitional probability for non-leaf nodes and prior class probability for leaf nodes remain the same as before (Table~\ref{tab:transitionprob}). 
What is different is the conditional probability of sample feature vector given its class. In the extended multi-modal model, sample feature in each class is assumed i.i.d. mixture Gaussian distribution, as shown in Equation~\ref{eq:featureclassprob2}, where $\boldsymbol{\mu}_{y_n}^i$ and $\boldsymbol{\Sigma}_{y_n}^i$ are the mean and covariance matrix of the $i$-th Gaussian component of feature vector $\mathbf{x}_n$ for class $y_n$ ($y_n=0,1$),  $\phi^i_{y_n}$ is the probability of a sample  in class $y_n$ belonging to the $i$th Gaussian component $\mathcal{N}(\boldsymbol{\mu}^i_{y_n},\boldsymbol{\Sigma}^i_{y_n})$, $K_{y_n}$ is the number of Gaussian components (modes) in the feature distribution of samples in class $y_n$. 

\begin{equation}\label{eq:featureclassprob2}
    P(\mathbf{x}_n|y_n)\sim \sum^{K_{y_n}}_{i=1} \phi^i_{y_n}\mathcal{N}(\boldsymbol{\mu}^i_{y_n},\boldsymbol{\Sigma}^i_{y_n}), y_n = 0, 1
\end{equation}

Based on the extended probabilistic formulation, we can use the same EM algorithm with message propagation for parameter learning. The entire set of parameters can be denoted as $\boldsymbol{\Theta}=\{\rho, \pi,\phi^i_c ,\boldsymbol{\mu}^i_c, \boldsymbol{\Sigma}^i_c|1\leq i \leq K_c, c=0,1 \}$. The EM algorithm involves iterations with each iteration consisting of two major steps: an E-step that calculates the marginal class posterior probabilities by message propagation based on old parameters, and an M-step that finds the optimal parameters to maximize the overall objective. We can use the same message propagation framework specified in Equation~\ref{eq:hmtforwardin}, Equation~\ref{eq:hmtforwardout}, Equation~\ref{eq:hmtbackwardin} and Equation~\ref{eq:hmtbackwardout}. The main difference is that the calculation of messages related to the factor $P(\mathbf{x}_n|y_n)$ is based on the mixture of Gaussian distribution instead of a single-modal Gaussian distribution, as specified in Equation~\ref{eq:featureclassprob2}. 

We can use the same framework in Equation~\ref{eq:unnormalizedMarginY}, Equation~\ref{eq:unnormalizedMarginYYp}, Equation~\ref{eq:normalizedMarginY} and Equation~\ref{eq:normalizedMarginYYp} to estimate the marginal posterior class probability. After the computation of marginal posterior distribution, we can update model parameters by maximizing the posterior expectation of log-likelihood (the maximization or M step in EM). Because the probabilistic formulation of feature distribution is extended to a mixture of Gaussian, the parameter update formula needs to be updated. The extended parameter update formulas are shown by equations below. The symbol $\boldsymbol{\Theta_0}$ represents the old parameter values in the previous iteration of the EM algorithm, which is used in the calculation of messages and estimated posterior class probabilities. These calculated terms are then used to update the parameters in the formulas below to get a new $\boldsymbol{\Theta}$ (the maximization step or M-step in the EM algorithm).

\begin{equation}\label{eq:updaterho2}\footnotesize
   \rho = \frac{\sum\limits_{n|\mathcal{P}_n\neq\emptyset}{\sum\limits_{y_n}\sum\limits_{y_{\mathcal{P}_n}}{y_{\mathcal{P}_n}(1 - y_n)P(y_n, y_{\mathcal{P}_n}|\mathbf{X}, \boldsymbol{\Theta_0})}}} 
 {\sum\limits_{n|\mathcal{P}_n\neq\emptyset}{\sum\limits_{y_n}\sum\limits_{y_{\mathcal{P}_n}}{y_{\mathcal{P}_n}P(y_n, y_{\mathcal{P}_n}|\mathbf{X}, \boldsymbol{\Theta_0})}}} 
\end{equation}
%parameter update

\begin{equation}\label{eq:updatepi2}\footnotesize
  \pi = \frac{\sum\limits_{n|\mathcal{P}_n=\emptyset}{\sum\limits_{y_n}{ y_n P(y_n|\mathbf{X}, \boldsymbol{\Theta_0})}}} {\sum\limits_{n|\mathcal{P}_n=\emptyset}{\sum\limits_{y_n}{P(y_n|\mathbf{X}, \boldsymbol{\Theta_0})}}}  
\end{equation}

\begin{equation}\label{eq:updatephi2}\footnotesize
  \phi^i_c = \frac{\sum\limits_{n \in \mathcal{O}} P(y_n = c|\mathbf{X},\boldsymbol{\Theta_0}) \gamma^i_c(x_n,\boldsymbol{\Theta_0})} {\sum\limits^{K_c}_{i= 1} \sum\limits_{n \in \mathcal{O}}  P(y_n = c|\mathbf{X},\boldsymbol{\Theta_0})\gamma^i_c(x_n,\boldsymbol{\Theta_0})},c = {0, 1}
\end{equation}

\begin{equation}\label{eq:updatemu2}\footnotesize
    \mu^i_c = \frac{\sum\limits_{n \in \mathcal{O}} \mathbf{x}_n P(y_n = c|\mathbf{X},\boldsymbol{\Theta_0}) \gamma^i_c(x_n,\boldsymbol{\Theta_0})} {\sum\limits_{n \in \mathcal{O}}  P(y_n = c|\mathbf{X},\boldsymbol{\Theta_0})\gamma^i_c(x_n,\boldsymbol{\Theta_0})},c = {0, 1}
\end{equation}

\begin{equation}\label{eq:updatesigma2}\footnotesize
    \Sigma_c^i = \frac{\sum\limits_{n \in \mathcal{O}} (\mathbf{x}_n - \boldsymbol{\mu}_c) (\mathbf{x}_n - \boldsymbol{\mu}_c)^T P(y_n = c|\mathbf{X},\boldsymbol{\Theta_0})
    \gamma^i_c(x_n,\boldsymbol{\Theta_0})} {\sum\limits_{n \in \mathcal{O}} P(y_n = c|\mathbf{X},\boldsymbol{\Theta_0})\gamma^i_c(x_n,\boldsymbol{\Theta_0})}, c = {0, 1},
\end{equation}

\begin{equation}\label{eq:gammaeq}\footnotesize
 \text{where} \quad \gamma^i_c(\mathbf{x}_n,\boldsymbol{\Theta_0}) = \frac{\phi^{i}_{0,c}\mathcal{N}(\mathbf{x}_n|\boldsymbol{\mu}^i_{0,c},\boldsymbol{\Sigma}^i_{0,c})}{\sum\limits^{K_c}_{i=1}\phi^{i}_{0,c}\mathcal{N}(\mathbf{x}_n|\boldsymbol{\mu}^i_{0,c},\boldsymbol{\Sigma}^i_{0,c}) }
    ,c = {0, 1},
\end{equation}

In the new parameter update formulas, the formulas for $\rho$ and $\pi$ are similar to the single-modal version. The main difference is related to parameters for feature distributions, such as $\phi^i_c$, $\mu^i_c$ and $\Sigma_c^i$ in Equation~\ref{eq:updatephi2}, Equation~\ref{eq:updatemu2}, and Equation~\ref{eq:updatesigma2}. The key difference is the additional weighting terms $\gamma^i_c(x_n,\boldsymbol{\Theta_0})$ in Equation~\ref{eq:gammaeq}. Intuitively, the intermediate variable $\gamma^i_c(x_n,\boldsymbol{\Theta_0})$ reflects the weight of a sample with feature $\mathbf{x}_n$ belong to the $i$-th Gaussian component of class $c$. For example, in the parameter update formula for $\mu^i_c$ (Equation~\ref{eq:updatemu2}), the updated mean for features in the $i$-th component of class $c$ is based on the feature vector of each sample, averaged by the sample's weight on the $i$-th component of class $c$ ($\gamma^i_c(x_n,\boldsymbol{\Theta_0})$). Similar interpretation can be given on the update formulas for $\phi^i_c$ and $\Sigma_c^i$. 

Another issue is the parameter initialization before the EM iteration, the initial parameters of $\phi^i_c$, $\mu^i_c$ and $\Sigma_c^i$ can be initialized based on a small set of training samples (pixels) with known labels. This can be done by an EM clustering for feature values in each class. The cluster centroids can be initialized by randomly select $K_c$ samples from the training samples in class $c$.

After parameter learning, we can use the same class inference algorithm based on message propagation (\emph{max-sum}~\cite{rabiner1989tutorial}). The calculation of messages is similar to before, except that the message related to $P(\mathbf{x}_n|y_n)$ are calculated based on the mixture of Gaussian distribution. 
\section{Experimental Evaluation}\label{sec:eval}

\subsection{Experiment Setup}
In this section, we compared our proposed approach with baseline methods in related works on real-world datasets. Evaluation candidate methods are listed below. Note that we did not include data imputation methods (e.g., filling in mean feature values) due to its low capability of handling very limited observations. Unless specified otherwise, we used default parameters in open source tools for baseline methods. Experiments were conducted on a Dell workstation with Intel(R) Xeon(R) CPU E5-2687w v4 @ 3.00GHz, 64GB main memory, and Windows 10. 

\begin{itemize}
    \item {\bf Label propagation with structure (LP-Structure)}: In the implementation of this baseline method, we used the maximum likelihood classifier (MLC) and gradient boosted model (GBM) respectively to pre-classify fully observed samples and then ran label propagation~\cite{wang2007label} on the topography tree structure. We named them as {\bf LP-Structure-MLC} and {\bf LP-Structure-GBM}. The initial classifiers were from R packages. 
    \item {\bf EM with i.i.d. assumption (EM-i.i.d.)}: In the implementation of this baseline method~\cite{ghahramani1994learning}, we treated missing features and unknown classes as latent variables and used the EM algorithm assuming that sample features follow i.i.d. Gaussian distribution in each class. Moreover, we assumed RGB (red, green, blue) features and elevation features are uncorrelated. 
    \item {\bf EM with structure}: This is our proposed approach. We treated unknown classes as latent variables and used the EM algorithm assuming that samples follow the topography tree dependency structure. The codes were implemented in C++. There are two configurations:  single-modal feature distribution ({\bf EM-Structure-Single}) and multi-modal feature distribution ({\bf EM-Structure-Multi}).
\end{itemize}

\emph{Data Description:} Our real-world datasets were collected from Kinston North Carolina and Grimesland North Carolina in Hurricane Matthew 2016. We used aerial imageries from NOAA National Geodetic Survey~\cite{ngs} with red, green, blue bands in a 2-meter spatial resolution and digital elevation map from the University of North Carolina Libraries~\cite{ncsudem}. The test region size was 1743 by 1349 in Kinston and 2757 by 3853 in Grimesland. The number of observation samples was 31,168 in Kinston and 237,312 in Grimesland.
The number of training and testing samples (pixels) are listed in Table~\ref{tab:data}. 

\begin{table}[h]\footnotesize
\centering
\caption{Dataset description}
\begin{tabular}{lcccc}
\hline
\multirow{2}{*}{Dataset} & \multicolumn{2}{c}{Training Set}  & \multicolumn{2}{c}{Testing Set} \\ 
&Dry & Flood & Dry & Flood \\ \hline
Matthew, Kinston  & 5,000 & 5,000 & 48,071 & 47,967 \\ \hline
Matthew, Grimesland  & 5,000 & 5,000 & 75,670 & 59,405 \\ \hline
\end{tabular}
\label{tab:data}
\end{table}

\emph{Evaluation Metrics}: For classification performance evaluation, we used precision, recall, and F-score. For computational performance evaluation, we measured the running time costs in seconds.

\subsection{Classification Performance Evaluation}
%for TC4
\begin{table}[h]\footnotesize
\centering
\caption{Comparison on Mathew, Kinston flood data}
\label{tab:comp1}
\begin{tabular}{cccccc}
\hline
Classifiers & Class & Prec. &Recall & F & Avg. F\\ \hline
\multirow{2}{*}{LP-Structure-GBM}&Dry&{0.91}&{0.56}&{0.69}&\multirow{2}{*}{0.74}\\ 
 &Flood&{0.68}&{0.94}&{0.79}&\\ \hline
\multirow{2}{*}{LP-Structure-MLC}&Dry&{0.86}&{0.55}&{0.67}&\multirow{2}{*}{0.72}\\ 
 &Flood&{0.67}&{0.91}&{0.77}&\\ \hline 
 \multirow{2}{*}{EM-i.i.d.}&Dry&{1.00}&{0.39}&{0.56}&\multirow{2}{*}{0.66}\\ 
 &Flood&{0.62}&{1.00}&{0.76}&\\ \hline 
\multirow{2}{*}{EM-Structure-Single}&Dry&{0.94}&{0.99}&{0.96}&\multirow{2}{*}{0.96}\\ 
 &Flood&{0.99}&{0.94}&{0.96}&\\ \hline  
\end{tabular}
\end{table}

%for TC1
\begin{table}[h]\footnotesize
\centering
\caption{Comparison on Mathew, Grimesland flood data}
\label{tab:comp2}
\begin{tabular}{cccccc}
\hline
Classifiers & Class & Prec. &Recall & F & Avg. F\\ \hline
\multirow{2}{*}{LP-Structure-GBM}&Dry&{0.81}&{0.60}&{0.69}&\multirow{2}{*}{0.70}\\ 
 &Flood&{0.61}&{0.82}&{0.70}&\\ \hline
\multirow{2}{*}{LP-Structure-MLC}&Dry&{0.90}&{0.75}&{0.82}&\multirow{2}{*}{0.81}\\ 
 &Flood&{0.73}&{0.90}&{0.81}&\\ \hline 
  \multirow{2}{*}{EM-i.i.d.}&Dry&{0.83}&{0.74}&{0.78}&\multirow{2}{*}{0.77}\\ 
 &Flood&{0.71}&{0.80}&{0.75}&\\ \hline 
\multirow{2}{*}{EM-Structure-Single}&Dry&{0.99}&{0.96}&{0.97}&\multirow{2}{*}{0.97}\\ 
 &Flood&{0.95}&{0.99}&{0.97}&\\ \hline  
\end{tabular}
\end{table}
We first compared different methods on their precision, recall, and F-score on the two real-world datasets. The results were summarized in Table~\ref{tab:comp1} and Table~\ref{tab:comp2} respectively. On the Kinston dataset, EM algorithm with the i.i.d. assumption performed the worst with an average F-score of 0.66. The reason was that this method was not able to utilize the spatial structural constraint between sample classes. Its training process only updated the parameter of Gaussian feature distribution in each class. When predicting the classes of samples with only elevation feature, the method used only the learned Gaussian distribution of elevation feature on each class without considering spatial structure based on elevation values. On the same dataset, label propagation after pre-classification with the GBM model and the maximum likelihood classifier slightly outperformed the EM algorithm with the i.i.d. assumption. The main reason was that label propagation on the topography tree (split tree) structure utilized the physical constraint between sample classes when inferring the classes of unobserved samples without RGB features. However, label propagation still showed significant errors, particularly in the low recall on the dry class. Through analyzing the predicted map, we observed that the label propagation algorithm was very sensitive to the pre-classified class labels on the observed samples in the test region. Errors in the pre-classification phase may propagation into unobserved samples (those without RGB feature values). In label propagation methods, once the errors were propagated to unobserved samples, they were hard to be reverted. This was different from the EM algorithm, which could update the probabilities in iterations. We did not report the results of label propagation on a grid graph structure (only considering spatial neighborhood structure without physics-aware constraint) due to poor results. Our model based on the EM algorithm assuming structural dependency between class labels performed the best with an average F-score of 0.96. The main reason was that our model could leverage the physical constraint to infer unobserved samples, and also could effectively update sample probabilities during iterations with the EM algorithm. In our model, we used training samples to initialize the parameters of the Gaussian distribution of sample features in each class. Based on the reasonable initial parameters, we can have a reasonable estimation of the posterior class probabilities of all samples in the test region. Based on the posterior class probabilities, the distribution parameters could be further updated. The representative training samples helped make sure that parameter iterations would converge in the right path.

Similar results were observed on the Grimesland dataset. In the label propagation method, pre-classification based on GBM performed worse than pre-classification based on MLC. The reason may be due to overfitting of GBM compared with MLC when predicting initial labels on the fully observed samples. The EM algorithm with the i.i.d. assumption performed slightly better on this dataset. The reason is likely that the final prediction of classes of the unobserved samples (with only elevation feature but without RGB features) was based on a slightly better fitted normal distribution. Our model showed the best performance with an F-score of 0.97.

\begin{figure}[h]
\centering
\subfigure[]{%
      \includegraphics[width=1.5in]{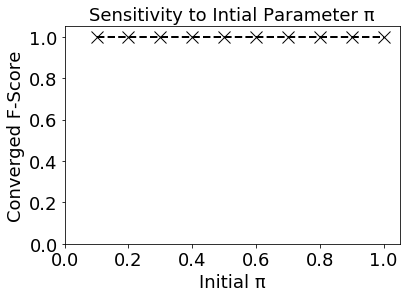}
}
\subfigure[]{%
       \includegraphics[width=1.5in]{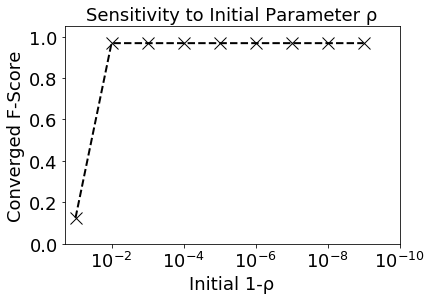}
 } 
\caption{Sensitivity of our model to initial parameters $\pi$ and $\rho$}
\label{fig:realSensitivity}
\end{figure}

\subsubsection{The effect of model initial parameters}
We also analyzed the sensitivity of our proposed model on different initial parameter settings. The parameters of $\boldsymbol{\mu}_c$ and $\boldsymbol{\Sigma}_c$ were estimated from training data, but parameters $\rho$ and $\pi$ were from user input. Since $\rho$ captured the transitional probability of a sample being flood given its parents were all flood, its value should be very high (close to 1) due to spatial autocorrelation. $\pi$ is the initial class prior probabilities for samples without parent nodes (local lowest location). We could set it close to $0.5$. We tested the sensitivity of our model to different initial values of $\rho$ and $\pi$ on the Kinston dataset. We first fixed $\rho$ as $0.999$ and varied the value of $\pi$ from $0.1$ to $0.9$. Then we fixed $\pi$ as $0.3$ and varied the value of $\rho$ from $0.9$. The results were shown in Figure~\ref{fig:realSensitivity}. We can see that the model was generally not very sensitive to the initial parameter values. For parameter $\rho$, as long as $1-\rho$ was smaller than $0.01$ ($\rho$ greater than $0.99$), the converged F-score was good. For parameter $\pi$, the results were consistently good for our model with an initial $\pi$ between $0.1$ to $0.9$. The main reason was that $\pi$ influenced only a small number of samples at the local lowest locations on the elevation map.

\begin{figure}[h]
\centering
\includegraphics[width=1.5in]{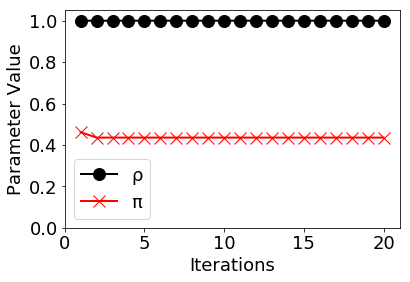}
\caption{Parameter iterations and convergence in our model}
\label{fig:realIterations}
\end{figure}

The parameter iterations of our model were shown in Figure~\ref{fig:realIterations}. The model converged fast with only 20 iterations. Due to the space limit, we only show the parameters of $\rho$ and $\pi$.

\subsection{Computational Performance Evaluation}
\begin{figure}[h]
\centering
\includegraphics[width=1.5in]{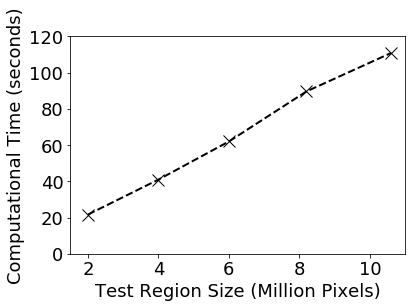}
\caption{Computational performance of out model on varying test region sizes}
\label{fig:timecost}
\end{figure}

We also evaluated the computational performance of our model learning and inference on different input data sizes. We used the Grimesland dataset to test the effect of different test region sizes. We varied the region size from around 2 million pixels to over 10 million pixels. The computational time costs of our model were shown in Figure~\ref{fig:timecost}. It can be seen that the time cost grows almost linearly with the size of the test region. This was because our learning and inference algorithms involve tree traversal operations with a linear time complexity on the tree size (the number of pixels on the test region). The model is computationally efficient. It classified around 10 million pixels in around 2 minutes. 

We further analyzed the time costs of different components in our model, including split tree construction, model parameter learning, and class inference. We analyzed the results on both datasets (same as the settings in Table~\ref{tab:comp1} and Table~\ref{tab:comp2}. Results showed that tree construction and class inference took less time than parameter learning. This was because the learning involves multiple iterations of message propagation (tree traversal operations). 
\begin{table}[h]
\centering
\caption{Time Costs Analysis of Our Model (seconds)}
\label{tab:costs}
\begin{tabular}{lccc}\hline
         & Kinston & Grimesland  \\ \hline
Tree construction & 3.2 & 8.39     \\\hline
Parameter learning & 25.74 & 86.79  \\\hline
Class inference    & 3.8      & 15.62     \\\hline
Total time        & 32.74 & 110.80  \\\hline
\end{tabular}
\end{table}

\subsection{Additional Comparison of EM Structured between Single-modal and Multi-modal}
This subsection provides additional evaluations on the comparison of our EM structured algorithms between the single-modal feature distribution (conference version) and multi-modal feature distribution (journal extension). In previous experiments, we found that our EM structured (single-modal) outperformed several baseline methods when feature distribution among training samples in each class is single-modal. In this experiment, we used the same two test regions and observed polygons as previous experiments (Table~\ref{tab:data}), but collected training samples that exhibit multi-modal feature distributions in each class. The new training dataset still contains 5,000 samples for dry class and 5,000 samples for flood class.

For both single-modal and multi-modal, we fixed the parameter $\pi$ (prior class probability for leaf nodes) as 0.5, $\rho$ as 0.999 (class transitional probability between a node and its parents) and the maximum number of iterations in EM as 40. We compared different candidate methods on their precision, recall, and F-score on the two test regions. 

%for TC4
\begin{table}[h]\footnotesize
\centering
\caption{Comparison on Mathew, Kinston flood data}
\label{tab:comp3}
\begin{tabular}{cccccc}
\hline
Classifiers & Class & Prec. &Recall & F & Avg. F\\ \hline
\multirow{2}{*}{LP-Structure-GBM}&Dry&{0.44}&{0.73}&{0.55}&\multirow{2}{*}{0.34}\\ 
 &Flood&{0.26}&{0.09}&{0.14}&\\
 \hline
%\multirow{2}{*}{LP-Structure-GBM}&Dry&{0.50}&{0.86}&{0.63}&\multirow{2}{*}{0.42}\\ 
% &Flood&{0.50}&{0.13}&{0.21}&\\ \hline
 \multirow{2}{*}{LP-Structure-MLC}&Dry&{0.85}&{0.56}&{0.67}&\multirow{2}{*}{0.72}\\ 
 &Flood&{0.67}&{0.90}&{0.77}&\\ \hline 
%\multirow{2}{*}{LP-Structure-MLC}&Dry&{0.76}&{0.54}&{0.63}&\multirow{2}{*}{0.68}\\ 
% &Flood&{0.64}&{0.82}&{0.72}&\\ \hline 
 \multirow{2}{*}{EM-i.i.d.}&Dry&{1.00}&{0.40}&{0.57}&\multirow{2}{*}{0.67}\\ 
 &Flood&{0.62}&{1.00}&{0.77}&\\ \hline 
\multirow{2}{*}{EM-Structure-Single}&Dry&{0.50}&{1.00}&{0.66}&\multirow{2}{*}{0.34}\\ 
 &Flood&{1.00}&{0.01}&{0.01}&\\ \hline 
  \multirow{2}{*}{EM-Structure-Multi}&Dry&{0.94}&{0.99}&{0.97}&\multirow{2}{*}{0.97}\\ 
 &Flood&{0.99}&{0.94}&{0.97}&\\ \hline  
\end{tabular}
\end{table}

%for TC2
\begin{table}[h]\footnotesize
\centering
\caption{Comparison on Mathew, Grimesland flood data}
\label{tab:comp4}
\begin{tabular}{cccccc}
\hline
Classifiers & Class & Prec. &Recall & F & Avg. F\\ \hline
\multirow{2}{*}{LP-Structure-GBM}&Dry&{0.74}&{0.53}&{0.62}&\multirow{2}{*}{0.71}\\ 
 &Flood&{0.75}&{0.88}&{0.81}&\\ \hline
%\multirow{2}{*}{LP-Structure-GBM}&Dry&{0.68}&{0.52}&{0.59}&\multirow{2}{*}{0.69}\\ 
 %&Flood&{0.74}&{0.85}&{0.79}&\\ \hline
 \multirow{2}{*}{LP-Structure-MLC}&Dry&{0.84}&{0.96}&{0.90}&\multirow{2}{*}{0.91}\\ 
 &Flood&{0.97}&{0.89}&{0.93}&\\ \hline
%\multirow{2}{*}{LP-Structure-MLC}&Dry&{0.80}&{0.95}&{0.87}&\multirow{2}{*}{0.89}\\ 
% &Flood&{0.96}&{0.86}&{0.91}&\\ \hline 
 \multirow{2}{*}{EM-i.i.d.}&Dry&{0.57}&{0.87}&{0.69}&\multirow{2}{*}{0.70}\\ 
 &Flood&{0.88}&{0.590}&{0.70}&\\ \hline 
\multirow{2}{*}{EM-Structure-Single}&Dry&{0.38}&{1.00}&{0.55}&\multirow{2}{*}{0.28}\\ 
 &Flood&{1.00}&{0.00}&{0.00}&\\ \hline 
 
 %\multirow{2}{*}{EM-Structure-Single}&Dry&{0.64}&{1.00}&{0.78}&\multirow{2}{*}{0.78}\\ 
 %&Flood&{1.00}&{0.65}&{0.78}&\\ \hline 
 
 \multirow{2}{*}{EM-Structure-Multi}&Dry&{0.91}&{0.96}&{0.94}&\multirow{2}{*}{0.95}\\ 
 &Flood&{0.97}&{0.94}&{0.96}&\\ \hline  
\end{tabular}
\end{table}

\begin{figure}[h]
\centering
\subfigure[Limited observation aerial imagery in Kinston NC]{%
      \includegraphics[width=1.5in]{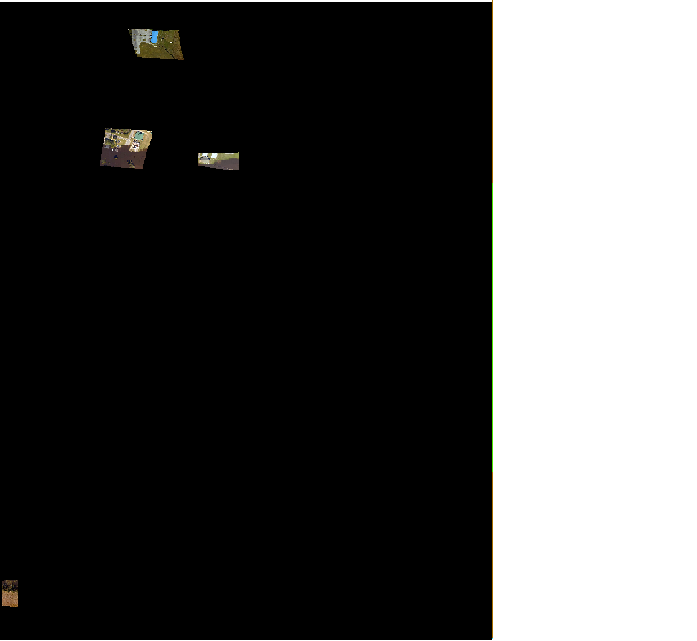}
}
\subfigure[Digital elevation model]{%
      \includegraphics[width=1.5in]{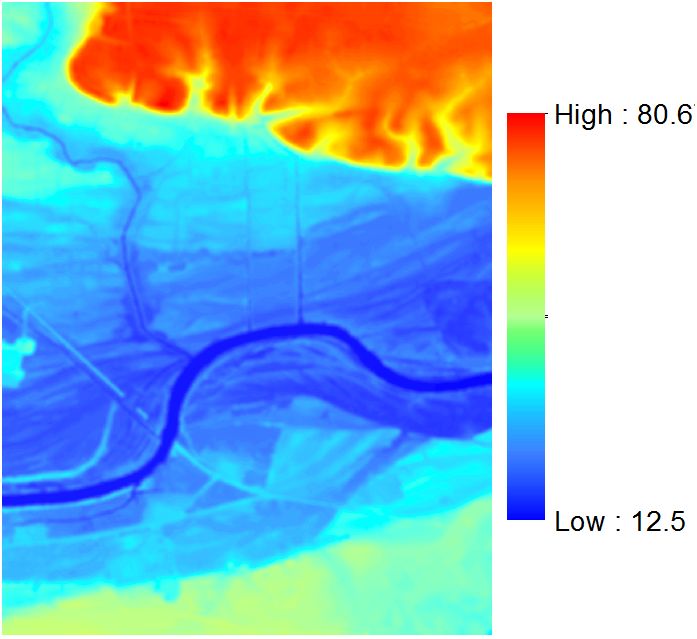}
}
% \subfigure[Flow direction]{%
%       \includegraphics[width=2.5in]{IJRS/figures/TC3/TC3FlowDirection.png}
% }
\subfigure[LP-Structure-GBM result]{%
      \includegraphics[width=1.5in]{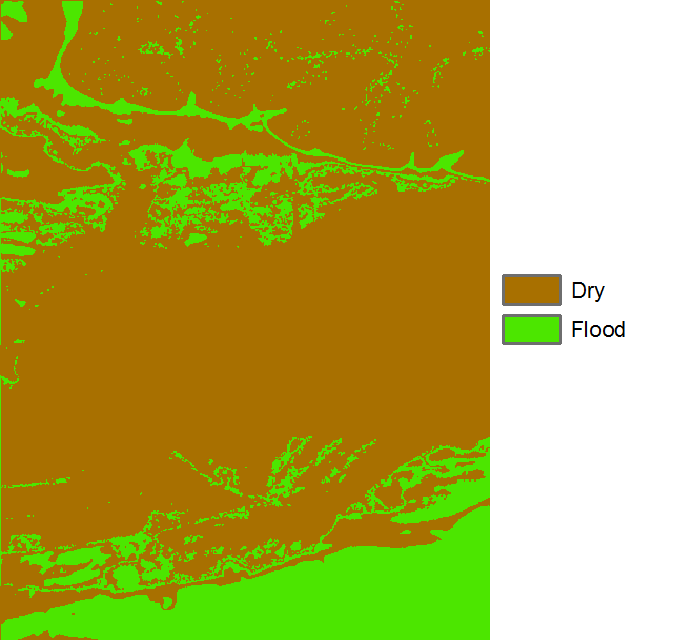}
}
\subfigure[LP-Strucrure-MLC result]{%
      \includegraphics[width=1.5in]{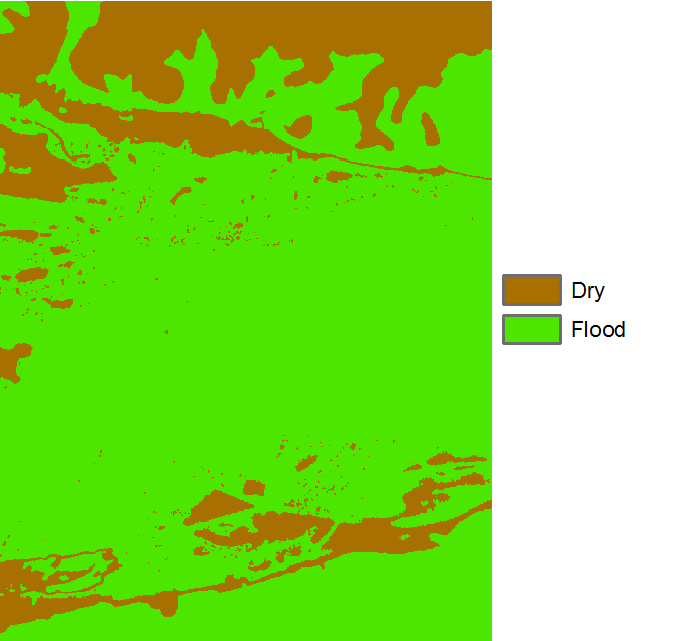}
}
\subfigure[EM-i.i.d. result]{%
      \includegraphics[width=1.5in]{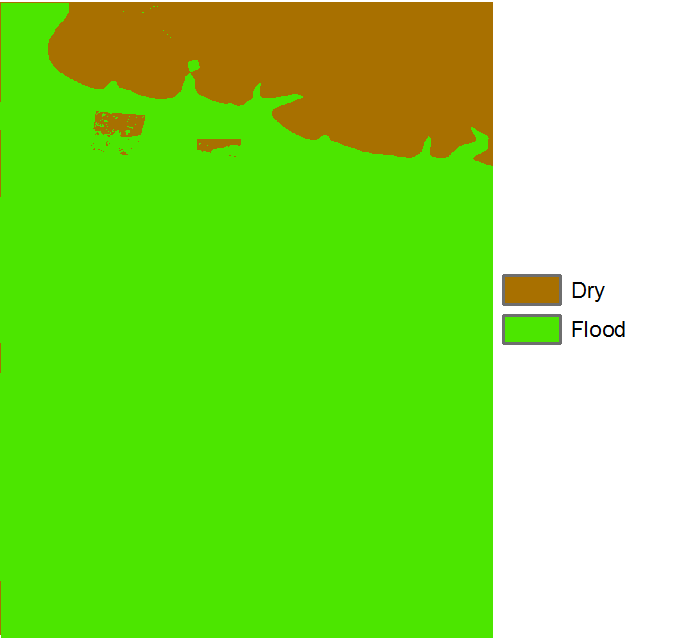}
}
\subfigure[EM-Structure-Single result]{%
      \includegraphics[width=1.5in]{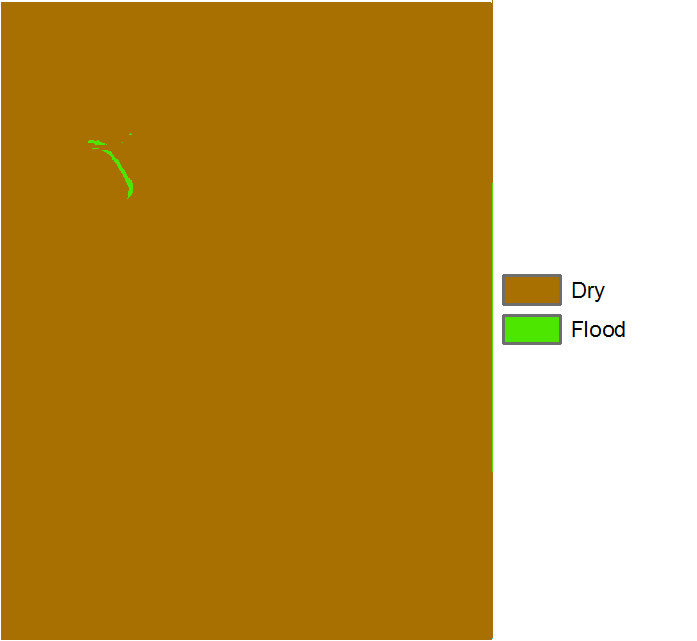}
}
\subfigure[EM-Structure-Multi result]{%
      \includegraphics[width=1.5in]{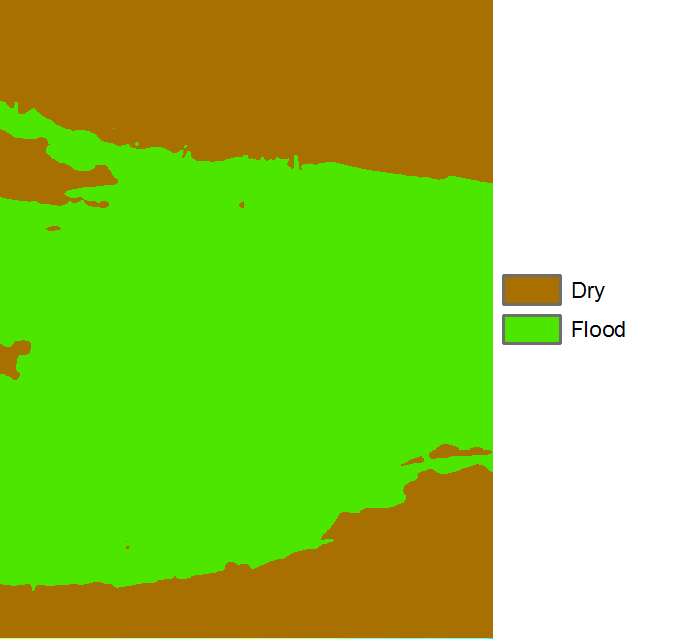}
}
\caption{Results on Matthew flood, Kinston, NC (flood in brown, dry in green, best viewed in color)}
\label{fig:map1}
\end{figure}

The new results were summarized in Table~\ref{tab:comp3} and Table~\ref{tab:comp4}. On Kinston flood data, we can see that the performance of the three baseline methods was poor (F-score below 0.72). Among these methods, GBM with label propagation (LP) was worse (F-score around 0.34), likely due to serious overfitting issues from the varying feature distributions between the training and test samples. The performance of the proposed EM-structure model with single-modal feature distribution (conference version) significantly degraded (F-score dropped to 0.34). The reason was probably that the feature values of training samples follow a multi-modal distribution, violating the original assumption that features are single-modal Gaussian in each class. Thus, the parameter iterations were not ineffective during the learning iterations on observed samples in the test region. In contrast, the EM-structured with multi-modal distribution was more robust with the best performance (F-score around 0.97). Similar results were seen in the Grimesland dataset. The three baseline methods generally performed poorly, except for the label propagation with the maximum likelihood method achieving an F-score of 0.91. The reason was likely that the maximum likelihood classifier was simple and less prone to overfitting in initial label prediction. The EM-structured model with a single modal performed poorly (F-score around 0.28) due to the wrong assumption on feature distribution. In contrast, the EM-structured model with a multi-modal assumption performed the best with an F-score around 0.95. 

We also visualize the predicted class maps on the Kinston dataset in Figure~\ref{fig:map1} for interpretation. The input limited feature observations on spectral pixels in the test region are shown in Figure~\ref{fig:map1}(a). The observed pixels cover parts of the boundary of the flood region in the test region. The digital elevation image that was used to construct a tree structure based on physical constraint is shown in Figure~\ref{fig:map1}(b). From the topography of the area, we can see that the test region has a flood plain surrounding a river channel that is spreading through the lower half of the image. The top and bottom areas of the image have higher elevation. The predictions of label propagation on top of GBM (LP-Structure-GBM) and MLC (LP-Structure-MLC) are in Figure~\ref{fig:map1}(c)-(d). We can see large areas of misclassification (e.g., the flood area in the middle of Figure~\ref{fig:map1}(c) is mistakenly predicted as dry, the dry area at the bottom of Figure~\ref{fig:map1} is mistakenly classified as flood). The reason is probably that initial class predictions by these models are noisy and these noisy labels further spread to bigger areas during label propagation. The EM i.i.d. algorithm also shows significant errors in the bottom part of the image. The reason is probably that the learned decision boundary on the elevation feature in the EM i.i.d. algorithm is inaccurate. The EM structured model with single-modal feature distribution performs poorly, classifying almost the entire area as dry. The reason is likely that the feature distribution in the model during parameter learning iterations is wrong, making the inferred classes largely wrong. In contrast, the EM structured model with multi-modal feature distribution identified the complete flood boundary.

\section{Conclusions and Future Work}\label{sec:con}
In this paper, we address the problem of spatial classification with limited feature observations. The problem is important in many applications where only a subset of sensors are deployed at certain regions or partial responses are collected in field surveys. Existing research on incomplete or missing data has limitations in assuming that incomplete feature observations only happen on a small subset of samples. Thus, these methods are insufficient for problems where the vast majority of samples have missing feature observations. To address this issue, we propose a new approach that incorporates physics-aware structural constraints into model representation. We propose efficient algorithms for model parameter learning and class inference. We also extend the model with multi-modal feature distribution based on the mixture Gaussian model. Evaluations on real-world hydrological applications show that our approach significantly outperforms several baseline methods in classification accuracy, and the proposed solution scales to a large data volume. Results also show that the proposed multi-modal extension is more robust than the single-modal version.

In future work, we plan to 
extend our proposed model to address other problems such as integrating noisy and incomplete observations such as volunteered geographic information (VGI). We also plan to explore the integration of deep learning framework with our approach.

\section*{Acknowledgement}
This material is based upon work supported by the NSF under Grant No. IIS 1850546 and the University Corporation for Atmospheric Research (UCAR).

\bibliographystyle{ACM-Reference-Format}
\bibliography{ref} 
\clearpage
\section*{Appendix}
\subsection{EM with structure (Our Approach)}

The parameters of hidden Markov tree include the mean and covariance matrix of sample features in each class, prior probability of leaf node classes, and class transition probability for non-leaf nodes. We denote the entire set of parameters as $\boldsymbol{\Theta}=\{\rho, \pi, \boldsymbol{\mu}_c, \boldsymbol{\Sigma}_c|c=0,1 \}$. We denote the set of samples with observed features as $\mathcal{O}$ and the set of samples with missing features as $\mathcal{M}$. Learning the set of parameters poses two major challenges: first, there exist unknown hidden class variables $\mathbf{Y}=[y_1,...,y_N]^T$, which are non-i.i.d.; second, the number of samples (nodes) is huge (up to hundreds of millions of pixels).

To address these challenges, we use the expectation-maximization (EM) algorithm and message (belief) propagation. The posterior expectation of log likelihood  is:
\begin{equation}\label{eq:postexpll}\scriptsize
\begin{split}
LL(\boldsymbol{\Theta}) & =\mathbb{E}_{\mathbf{Y}|\mathbf{X}_o,\boldsymbol{\Theta_0}}\log P(\mathbf{X}_o,\mathbf{Y}|\boldsymbol{\Theta})\\
&=\mathbb{E}_{\mathbf{Y}|\mathbf{X}_o,\boldsymbol{\Theta_0}}\log\left\{ \prod_{n\in \mathcal{O}} P(\mathbf{x}_n|y_n,\boldsymbol{\Theta}) \prod_{n=1}^NP(y_n|y_{k\in\mathcal{P}_n},\boldsymbol{\Theta})\right\}\\
&=\sum\limits_{\mathbf{Y}}{ P(\mathbf{Y}|\mathbf{X}_o,\boldsymbol{\Theta_0})}\\
&\quad\quad{\left\{\sum_{n\in \mathcal{O}}\log{P(\mathbf{x}_n|y_n,\boldsymbol{\Theta})}+\sum_{n=1}^{N}\log{P(y_n|y_{k\in\mathcal{P}_n},\boldsymbol{\Theta})}\right\}}\\
& =\sum_{n\in \mathcal{O}}\sum_{y_n}P({y}_n|\mathbf{X}_o,\boldsymbol{\Theta_0})\log{P(\mathbf{x}_n|y_n,\boldsymbol{\Theta})}\\
&\quad\quad+\sum_{n=1}^{N}~\sum_{y_n,y_{k\in\mathcal{P}_n}}P(y_n,y_{k\in\mathcal{P}_n}|\mathbf{X}_o,\boldsymbol{\Theta_0})\log{P(y_n|y_{k\in\mathcal{P}_n},\boldsymbol{\Theta})} \\
\end{split}
\end{equation}
Note that for leaf node, $\mathcal{P}_n=\emptyset$, and the last term in the last line of above equation is degraded,
$P(y_n,y_{k\in\mathcal{P}_n}|\mathbf{X},\boldsymbol{\Theta_0})\log{P(y_n|y_{k\in\mathcal{P}_n},\boldsymbol{\Theta})}=P(y_n|\mathbf{X},\boldsymbol{\Theta_0})\log{P(y_n|\boldsymbol{\Theta})}$.

In our model, we assume that the sample features follow Gaussian distribution in each class, then
\begin{equation}\label{eq:emgaudis}\scriptsize
    \begin{split}
        P(\mathbf{x}_n|y_n = c,\boldsymbol{\Theta}) = \frac{1}{\sqrt{2\pi|\Sigma_c|}}\exp\{-\frac{1}{2}(\mathbf{x}_n-\mu_c)^T|\Sigma_c|^{-1}(\mathbf{x}_n-\mu_c) \}, c = {0, 1}
    \end{split}
\end{equation}

After computation of marginal posterior distribution with forward and backward message propagation, we can put equation(9), (10) and equation(16) to the posterior expectation of log likelihood and update model parameters by maximizing it. Then it is straightforward to get the equation(11), (12), (13), and (14). 

\emph{Class inference}: After learning model parameters, we can infer hidden class variables by maximizing the overall probability. We use a dynamic programming based on method \emph{max-sum}. The process is similar to the sum and product algorithm. In message propagation, we use max operation and memorize the optimal variable values. 

\subsection{EM with i.i.d assumption}
For EM with i.i.d assumption, we do not consider the physics-aware structral constraint and assume the sample features follow i.i.d Gausian distribution. Moreover we also assume the sample non-spatial features and elevation features are uncorrelated. 

In our model, the entire set of parameters is $\boldsymbol{\Theta}=\{\rho, \pi, \boldsymbol{\mu}_c, \boldsymbol{\Sigma}_c|c=0,1 \}$. To learn these parameters, we first combine the non-spatial explanatory feature matrix and the elevation feature as the feature vector for each sample having observation and use the feature vector to learn parameters. Here we define the class of samples with complete explanatory features are noted as $\{\mathbf{y}_n|n\in\mathcal{O}\}$. The corresponding class matrix is noted as $\mathbf{Y_o}$. The expected log likelihood of observed sample is as below, 
\begin{equation}\label{eq:empostexpll}\scriptsize
    \begin{split}
    LL(\boldsymbol{\Theta}) & =\mathbb{E}_{\mathbf{Y}_o|\mathbf{X}_o,\boldsymbol{\Theta_0}}\log P(\mathbf{X}_o,\mathbf{Y}_o|\boldsymbol{\Theta})\\
    &=\mathbb{E}_{\mathbf{Y}_o|\mathbf{X}_o,\boldsymbol{\Theta_0}}\log \prod_{n\in \mathcal{O}} P(\mathbf{x}_n,y_n|\boldsymbol{\Theta})  \\
    % &=\int_\mathbf{Y}d\mathbf{Y}\\
    &=\sum\limits_{\mathbf{Y}_o} P(\mathbf{Y}_o|\mathbf{X}_o,\boldsymbol{\Theta_0}){\sum_{n\in \mathcal{O}}\log{P(\mathbf{x}_n,y_n|\boldsymbol{\Theta})}}\\
    & =\sum_{n\in \mathcal{O}}\sum_{y_n}P({y}_n|\mathbf{X}_o,\boldsymbol{\Theta_0})\log{P(\mathbf{x}_n,y_n|\boldsymbol{\Theta})} \\
    & =\sum_{n\in \mathcal{O}}\sum_{y_n}P({y}_n|\mathbf{x}_n,\boldsymbol{\Theta_0})\log{P(\mathbf{x}_n,y_n|\boldsymbol{\Theta})} \\
    \end{split}
\end{equation}

We assume that the sample features follow i.i.d Gaussian distribution in each class. Moreover, the non-spatial features and elevation feature are uncorrelated. The prior distribution of each class is uniform. Then the class prior ditribution $P(y_n = c|\boldsymbol{\Theta})$,  feature distribution $P(\mathbf{x}_n|y_n,\boldsymbol{\Theta})$ and joint distribution of feature and class $P(\mathbf{x}_n,y_n = c|\boldsymbol{\Theta})$  is as below,
\begin{equation}\label{eq:empriordis}\scriptsize
    P(y_n = c|\boldsymbol{\Theta}) = \pi_c, c = {0, 1}
\end{equation}

\begin{equation}\label{eq:emgaudis}\scriptsize
    \begin{split}
        P(\mathbf{x}_n|y_n = c,\boldsymbol{\Theta}) = \frac{1}{\sqrt{2\pi|\Sigma_c|}}\exp\{-\frac{1}{2}(\mathbf{x}_n-\mu_c)|\Sigma_c|^{-1}((\mathbf{x}_n-\mu_c))^T \}, c = {0, 1}
    \end{split}
\end{equation}
\begin{equation}\label{eq:emgaudis}\scriptsize
    \begin{split}
        P(\mathbf{x}_n,y_n = c|\boldsymbol{\Theta}) &=P(y_n = c|\boldsymbol{\Theta}) P(\mathbf{x}_n|y_n = c,\boldsymbol{\Theta})\\ &=\frac{\pi_c}{\sqrt{2\pi|\Sigma_c|}}\exp\{-\frac{1}{2}(\mathbf{x}_n-\mu_c)|\Sigma_c|^{-1}((\mathbf{x}_n-\mu_c))^T \}, c = {0, 1}
    \end{split}
\end{equation}
The class posterior distribution of class
$P({y}_n|\mathbf{X}_o,\boldsymbol{\Theta_0})$ is

\begin{equation}\label{eq:emposteriordis}\scriptsize
    \begin{split}
        P({y}_n = c|\mathbf{x}_n,\boldsymbol{\Theta_0}) = \frac{P(\mathbf{x}_n|y_n = c,\boldsymbol{\Theta_0})P(y_n = c)}{\sum_{y_n=0}^1P(\mathbf{x}_n|y_n,\boldsymbol{\Theta_0})P(y_n)}, c = {0, 1}
    \end{split}
\end{equation}
 
Taking the above into the posterior expectation of log likelihood into (17), we can easily get the following parameter to update formulas.
\begin{equation}\label{eq:priorupdate}\scriptsize
    \pi_c=\frac{ \sum_{n\in \mathcal{O}}P(y_n = c|\mathbf{x}_n,\boldsymbol{\Theta_0} )}{ \sum_{c=0}^1\sum_{n\in \mathcal{O}}P(y_n = c|\mathbf{x}_n,\boldsymbol{\Theta_0}) }
\end{equation}

\begin{equation}\label{eq:muupdate}\scriptsize
        \mu_c=\frac{ \sum_{n\in \mathcal{O}}P(y_n = c|\mathbf{x}_n,\boldsymbol{\Theta_0} )\mathbf{x}_n}{ \sum_{n\in \mathcal{O}}P(y_n = c|\mathbf{x}_n,\boldsymbol{\Theta_0}) }
\end{equation}

\begin{equation}\label{eq:sigmaupdate}\scriptsize
        \Sigma_c=\frac{ \sum_{n\in \mathcal{O}}P(y_n = c|\mathbf{x}_n,\boldsymbol{\Theta_0} )(\mathbf{x}_n - \mu_c)(\mathbf{x}_n - \mu_c)^T}{ \sum_{n\in \mathcal{O}}P(y_n = c|\mathbf{x}_n,\boldsymbol{\Theta_0}) }
\end{equation}

After learning model parameters, we can infer hidden class variables by maximizing the log likelihood.

\begin{equation}\label{eq:jointdis}\scriptsize
    \begin{split}
        \log P(\mathbf{X},\mathbf{Y}) &= \log \prod_{n\in \mathcal{O}}P(x_n|y_n)P(y_n)\prod_{n\in \mathcal{M}}P(x_n|y_n)P(y_n)\\ 
        &= \sum_{n\in \mathcal{O}}\log P(x_n|y_n)P(y_n) + \sum_{n\in \mathcal{M}}\log P(x_n|y_n)P(y_n)
    \end{split}
\end{equation}

To maximize the total log likelihood, we maximize each term of equation(25). For each sample n, we choose the class c that has higher probability of $P(x_n|y_n=c)P(y_n=c)$. The difference of the two parts of equation(25) is for the sample $n\in \mathcal{O}$ , $P(x_n|y_n)$ is multivariate normal distribution,  while for the sample $n\in \mathcal{M}$, $x_n$ only contains elevation feature, so the classification is based the normal distribution of elevation.

\end{document}